\documentclass[10pt,twocolumn,letterpaper, pagenumbers]{article}

\usepackage{cvpr}              

\usepackage[dvipsnames]{xcolor}

\usepackage{bm}
\newcommand{\realR}{\mathbb{R}}

\newcommand{\fiveDfiveC}{5^\circ5\text{cm}}

\newcommand{\fiveDtwoC}{5^\circ2\text{cm}}
\newcommand{\tenDtwoC}{10^\circ2\text{cm}}
\newcommand{\tenDfiveC}{10^\circ5\text{cm}}
\newcommand{\IoUSevenFive}{\text{IoU}_{75}}

\newcommand{\IoUFifty}{\text{IoU}_{50}}

\makeatletter
\newcommand\footnoteref[1]{\protected@xdef\@thefnmark{\ref{#1}}\@footnotemark}
\makeatother

\newcommand{\initR}{\mathbf{R}_{0}}
\newcommand{\initT}{\mathbf{t}_{0}}
\newcommand{\initS}{\mathbf{s}_{0}}
\newcommand{\deltaR}{\Delta\mathbf{R}}
\newcommand{\deltaT}{\Delta\mathbf{t}}
\newcommand{\deltaS}{\Delta\mathbf{s}}
\newcommand{\observedPoints}{\mathcal{O}}
\newcommand{\priorPoints}{\mathcal{P}}

\newcommand{\RFeature}{f_r}
\newcommand{\TSFeature}{f_\text{ts}}
\newcommand{\RMatrix}{M_r}
\newcommand{\TSMatrix}{M_\text{ts}}
\newcommand{\observedRFeature}{f^{O}_r}
\newcommand{\observedTSFeature}{f^{O}_\text{ts}}

\newcommand{\priorRMatrix}{M^{P}_r}
\newcommand{\priorTSMatrix}{M^{P}_\text{ts}}

\newcommand{\mixedRFeature}{f^\text{OP}_r}
\newcommand{\mixedTFeature}{f^\text{OP}_\text{t}}
\newcommand{\mixedSFeature}{f^\text{OP}_\text{s}}

\newcommand{\focalizedObservedPoints}{\hat{\mathcal{O}}}
\newcommand{\paperName}{GeoReF}
\newcommand{\focalizedPriorPoints}{\hat{\mathcal{P}}}
\newcommand{\degree}{^{\circ}}
\newcommand{\cm}{\text{cm}}

\newcommand{\diag}{\text{diag}}

\usepackage[accsupp]{axessibility}
\definecolor{cvprblue}{rgb}{0.21,0.49,0.74}
\usepackage[pagebackref,breaklinks,colorlinks,citecolor=cvprblue]{hyperref}
\usepackage{multirow}
\usepackage{bm}
\usepackage{adjustbox}

\definecolor{mydarkgreen}{RGB}{0, 128, 0}
\definecolor{mydarkred}{RGB}{139, 0, 0}

\def\paperID{13954} 
\def\confName{CVPR}
\def\confYear{2024}

\title{GeoReF: Geometric Alignment Across Shape Variation for Category-level Object Pose Refinement}

\author{
Linfang Zheng$^{1,3}$ \and Tze Ho Elden Tse\thanks{Equal contribution, order by dice rolling.} $^{\ 3}$\and Chen Wang$^{*\ 1,2}$ \and Yinghan Sun$^{1}$ \and Hua Chen$^{1}$ \and Ale\v{s} Leonardis$^{3}$ \and Wei Zhang\thanks{The corresponding author.} $^{\ 1}$ \and Hyung Jin Chang$^{3}$ \and\\$^{1}$Shenzhen Key Laboratory of Control Theory and Intelligent Systems, School of System Design and \\Intelligent Manufacturing, Southern University of Science and Technology, China\\
$^{2}$Department of Computer Science, the University of Hong Kong, China\\
$^{3}$School of Computer Science, University of Birmingham, UK\\
{\tt\small$\{$\text{lxz948,} \text{txt994}$\}$@student.bham.ac.uk, cwang5@cs.hku.hk, sunyh2021@mail.sustech.edu.cn}\\
{\tt\small $\{$chenh6,zhangw3$\}$@sustech.edu.cn,$\{$a.leonadis,h.j.chang$\}$@bham.ac.uk}
}

\begin{document}
\maketitle
\begin{abstract}
Object pose refinement is essential for robust object pose estimation. Previous work has made significant progress towards instance-level object pose refinement. Yet, category-level pose refinement is a more challenging problem due to large shape variations within a category and the discrepancies between the target object and the shape prior. To address these challenges, we introduce a novel architecture for category-level object pose refinement. Our approach integrates an HS-layer and learnable affine transformations, which aims to enhance the extraction and alignment of geometric information. Additionally, we introduce a cross-cloud transformation mechanism that efficiently merges diverse data sources. Finally, we push the limits of our model by incorporating the shape prior information for translation and size error prediction. We conducted extensive experiments to demonstrate the effectiveness of the proposed framework. Through extensive quantitative experiments, we demonstrate significant improvement over the baseline method by a large margin across all metrics.
\footnote{Project page: \href{https://lynne-zheng-linfang.github.io/georef.github.io}{https://lynne-zheng-linfang.github.io/georef.github.io}} 
\end{abstract}

\vspace{-0.2cm}
\section{Introduction}
\label{sec:intro}
\begin{figure}
\centering
\includegraphics[width=0.9\linewidth, trim = 20 10 130 50 , clip]{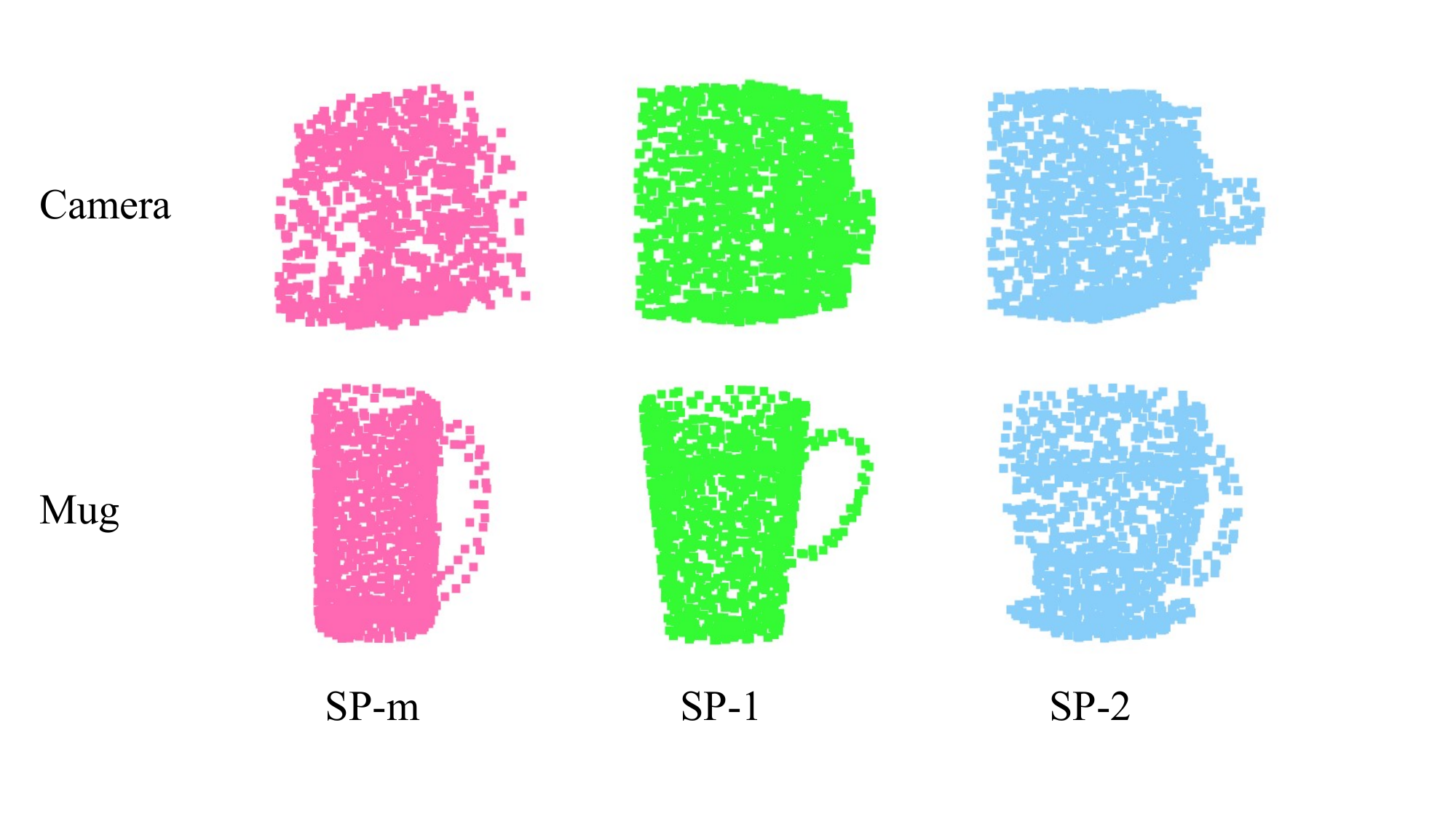}
\vspace{-6mm}
\caption{\footnotesize 
\textbf{Illustration of the shape variation.} \textit{SP-m} represents the category's mean shape, \textit{SP-1} and \textit{SP-2} represents the randomly sampled object shapes from the CAMERA25 training set. 
}

\label{fig:teaser}
\vspace{-12px}
\end{figure}

Understanding an object's pose is crucial for a wide range of real-world applications, including robotic manipulation~\cite{Wen-RSS-22, Kappler2018RAL, Morgan-RSS-21, Zhang2023ICRA}, augmented reality~\cite{Application_Augmented_Reality_2, AR_application_manufacturing}, and autonomous driving~\cite{augmented_reality_application_2021, autonomous_driving_application_2017}. Significant progress has been made for object pose estimation~\cite{TexPose_Chen_2023_CVPR, Densefusion_Wang_2019_CVPR, DPOD_2019_ICCV, EPOS_2020_CVPR, RigidityAwareDetection_Hai_2023_CVPR, PoseStatisticalGuarantees_Yang_2023_CVPR} and pose refinement~\cite{DeepIM_2018_ECCV, iwase2021repose, segal2009RSSgeneralized, SE3_TrackNet_2020_IROS, CRT6D_Castro_2023_WACV, Trans6D_Zhang_ECCV_Workshop_2022} using the object's CAD model. Despite the promising performance, the reliance on accurate instance-level CAD models limits their generalizability to everyday objects. Category-level methods~\cite{HSPose_2023_CVPR, SSP-Pose_IROS_22, SAR-Net_Lin_2022_CVPR, Self-DPDN_ECCV_2022, DualPoseNet_iccv_2021, CATRE_eccv_2022, StereoPose_ICRA_2023} is therefore been proposed to overcome this limitation. The objective of this line of work focuses on estimating object poses within a category given category-level shape priors. As a result, they face unique challenges as there exist diverse shape variations in each object category. We illustrate these shape variations in Fig.~\ref{fig:teaser}.


Recently, there have been remarkable advancements in category-level object pose estimation~\cite{HSPose_2023_CVPR, GPV-Pose_cvpr_2022, RBP-Pose_ECCV_2022, FSNET_2021_CVPR, Self-DPDN_ECCV_2022}, primarily due to effective utilization of geometric information through 3D graph convolution~\cite{3dgcn_lin_CVPR_2020}. In applications that require high precision, it is common to employ an object pose refinement procedure in conjunction with pose estimation. This involves an initial pose estimation algorithm determining the object pose, followed by a refinement step to further enhance the accuracy of the initially estimated pose by predicting and correcting its error. However, while instance-level object pose refinement has been extensively studied, category-level pose refinement remained unexplored until the introduction of CATRE~\cite{CATRE_eccv_2022}. By leveraging initial object pose and size estimations, CATRE achieves category-level pose refinement by iteratively aligning the observed target object point cloud with the category-level shape prior. This pipeline is shown to be effective by improving the accuracy of the initial pose and size estimations.

While CATRE has proven to be effective in many scenarios, it is limited by the reliance on the PointNet~\cite{Pointnet_2017_CVPR} encoder, which is primarily designed for classification and segmentation tasks. This design choice limits its ability to capture essential and fine-grained geometric relationships for accurate pose estimation and refinement. This ability is particularly important in category-level pose refinement as there exists diverse shape variations between inputs. Consequently, CATRE obtains suboptimal performance by direct application of 3D graph convolution. In addition, as CATRE treats the point cloud and the shape prior features separately until a later stage of the network, they potentially miss out on the benefits of integrating these features earlier. Moreover, their approach does not incorporate shape prior information into the translation and scale estimation module, which presents another area for potential improvement.

In this paper, we introduce a novel architecture for category-level object pose refinement which aims to address the limitations mentioned above. To better extract both local and global geometric information, we incorporate an HS layer into our feature extraction process. We apply learnable affine transformations to the features to address the geometric discrepancies between the observed point cloud and the shape prior. This enables the network to align these features more effectively. In addition, we propose a cross-cloud transformation mechanism that is specifically designed to enhance the merging of information between the observed point clouds and the shape prior. This mechanism enables more efficient integration of information between the two sources. Finally, we push the limit of our model by incorporating shape prior information to more accurately predict errors in translation and size estimation.

Our extensive experimental results on two category-level object pose datasets demonstrate that our proposed model to be effective in addressing the problem of shape variations in category-level object pose refinement, and consequently outperforms the state-of-the-art significantly. To the best of our knowledge, our proposed method is the first to successfully address the shape variation issue which is common in category-level pose refinement. Specifically, to enable graph convolution to be effective in capturing geometric relationships between different shapes, we propose an adaptive affine transformation matrix that aligns the observed point clouds and the shape prior. Additionally, the proposed cross-cloud transformation mechanism effectively fuses features from different input point clouds and brings further performance improvements. 

Our contributions are as follows:
\begin{itemize}
    \item {We introduce a novel architecture to specifically address the shape variations issue in category-level object pose refinement. Our proposed method results in consistent performance gain and exhibits better generalization ability.}
    \item We propose a unique cross-cloud transformation mechanism which efficiently merges diverse information from observed point clouds and shape priors.
    \item {We conduct extensive experiments on two category-level object pose datasets to validate our proposed method. On the REAL275 dataset, our method significantly outperforms SPD by $39.1\%$ increase in the $\fiveDfiveC$ metric. Additionally, we achieve $10.5\%$ improvement in the $\tenDtwoC$ metric over the state-of-the-art method, CATRE.}
\end{itemize}

\section{Related Work}
\paragraph{Instance-level object pose estimation and refinement.} {Instance-level approaches estimate the pose of the target object given known 3D CAD models. They can be briefly divided into correspondence matching methods and template matching methods.} Correspondence matching methods~\cite{SLAM6D_2022_CVPR, BB8_Rad_2017_ICCV, G2L_Net, PVN3D_CVPR_2019, Tremblay_2018_Bbox_Corners, ZebraPose_2022_CVPR, DPOD_2019_ICCV, OneShot_2d3d_2022_CVPR, Epro-PnP_2022_CVPR, SurfEmb_2022_CVPR, SSS_6D_2018_CVPR, DFTr_Zhou_2023_ICCV} matches the outstanding features of the observed object images with its model. Template matching methods~\cite{AAE_perspective, zheng2022TPAE, OSOP_2022_CVPR, OVE6D_2022_CVPR, MP-Encoder_CVPR_2020, Hinterstoisser_2016_PPF, Vidal_matching_PPF_2018, Temp6D_CVPR_22} compares the images or extracted features with the pre-generated templates. As the initial pose estimates can be noisy to various factors such as occlusions, object pose refinement~\cite{DeepIM_2018_ECCV, CosyPose_ECCV_2020, SE3_TrackNet_2020_IROS} is shown to be useful in improving the performance of instance-level methods. Even though they achieved impressive over the target object, the reliance on object CAD models limited their generalizability for handling everyday objects. {In this paper, we consider a more challenging problem setting where only the category-level shape prior is provided.}

\vspace{-2mm}
\paragraph{Category-level object pose estimation and refinement.} {Both tasks} mainly focus on addressing the shape variation between the objects. The pioneering work NOCS~\cite{NOCS_cvpr_2019} tackles the shape discrepancy by recovering the normalized visible shape of the target object and achieving the pose by point cloud matching. A series of methods extend this structure by leveraging different information such as domain adaptation~\cite{UDA-COPE_2022_CVPR}, different reconstruction space~\cite{CASS_2020_CVPR}, shape prior~\cite{SPD_eccv_2020, UDA-COPE_2022_CVPR, CASS_2020_CVPR, ShaPO_ECCV_2022}, and structural similarities~\cite{SGPA_iccv_2021, Self-DPDN_ECCV_2022}. However, this line of work is often limited in speed due to the iterative point matching. Another {series of} work starts with FS-Net~\cite{FSNET_2021_CVPR}, which adopts 3D graph convolution (3D-GC)~\cite{3dgcn_lin_CVPR_2020} to obtain geometric sensitivity. Due to its effectiveness and real-time performance, graph convolution is widely adopted in recent methods with an enhancement in directions including loss function~\cite{GPV-Pose_cvpr_2022}, bounding box voting~\cite{RBP-Pose_ECCV_2022}, and shape deformation~\cite{SSP-Pose_IROS_22}. HS-Pose~\cite{HSPose_2023_CVPR} extends the geometric feature extraction from local to global, which enhances the capability to handle objects with complex shapes. The research on category-level refinement {began} recently with the proposal of CATRE~\cite{CATRE_eccv_2022}. It introduced an effective pipeline that leverages shape priors and a focalization strategy for pose refinement and effectively improves the initial pose estimations. In this paper, we extend the CATRE and tackle the geometric variation issue within the framework of category-level pose refinement.

\begin{figure*}[!t]
\centering
\includegraphics[width=0.98\linewidth, trim = 00 0 0 0, clip]{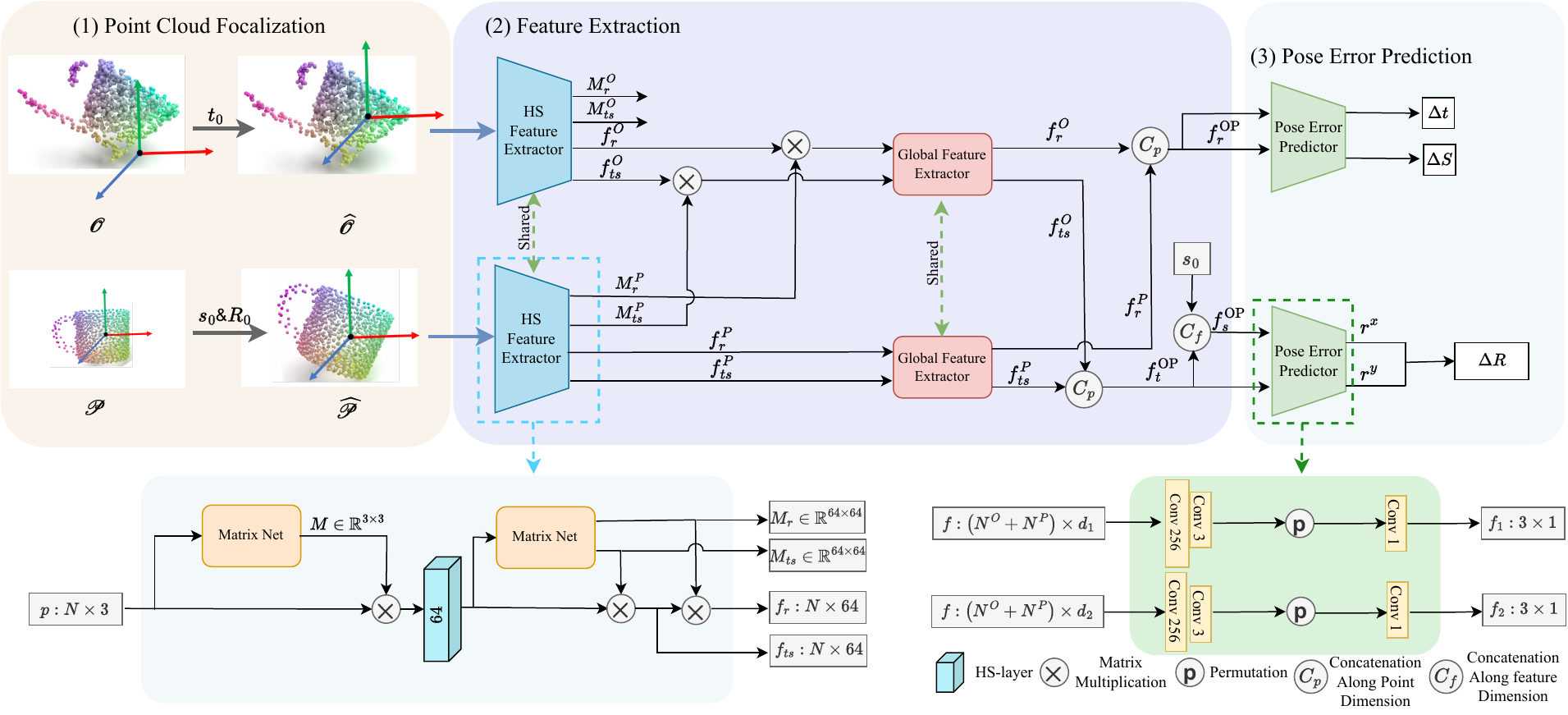}
\vspace{-10px}
\caption{\footnotesize \textbf{Overall structure of the proposed method.} Our object pose refinement structure contains three main modules. Given the shape prior point cloud, the target object's observed point cloud, and the initial estimation, we first apply point cloud focalization on the input point clouds using the initial estimation. The focalized point clouds then go through a geometric-based feature extraction encoder to obtain geometric structural features. The extracted features are then fed into two branches for rotation error estimation, translation error, and size error estimation. Within the HS Feature Extractor, the Matrix Net models output the learnable affine transformations (LATs) for adaptive point and feature adjustment. The left output of the Matrix Net adjusts the input point clouds, while the right Matrix Net model outputs two affine transformations for adjusting the rotation features, and the translation and size features.} 
\label{fig:framework}
\vspace{-12px}
\end{figure*} 
\section{Methodology}

\subsection{Problem Formulation}
In this paper, we tackle the problem of category-level object pose refinement. Given the initial pose and size estimation ($\initR$, $\initT$, $\initS$), along with the observed point cloud $\observedPoints \in \realR^{N^O\times3}$ and the shape prior $\priorPoints \in \realR^{N^P\times 3}$, we aim to predict the estimation error $(\deltaR$, $\deltaT$, $\deltaS)$ between the initial estimations and the ground truths. The pose refinement algorithm $\phi$ can be described as:
\begin{equation}
    (\deltaR, \deltaT, \deltaS) = \phi( \initR, \initT, \initS, \observedPoints, \priorPoints).
\end{equation}
This pose refinement algorithm $\phi$ can be applied iteratively to improve the refinement performance.

\subsection{Preliminaries}
Our proposed category-level object pose refinement framework builds upon two previous works, CATRE~\cite{CATRE_eccv_2022} and HS-layer~\cite{HSPose_2023_CVPR}, which we briefly review them in the following.

\vspace{-1mm}
\paragraph{CATRE.}
CATRE is the first framework that considers the problem of category-level pose refinement. It predicts the error between the ground truth and the estimated poses by aligning the input point clouds and the categorical shape priors. Specifically, the network architecture of CATRE consists of four components: a) point cloud focalization, b) shared encoder, c) rotation prediction, and d) translation and size prediction. In point clouds focalization, the observed point clouds $\observedPoints$ and the shape prior $\priorPoints$ are first aligned with the initial pose and size estimation $[\initR, \initT, \initS]$:
\vspace{-1mm}
\begin{equation}
    \begin{split}
        \focalizedObservedPoints = \{\hat{o}_i | \hat{o}_i = o_i - \initT, o_i \in \observedPoints\}, \\
        \focalizedPriorPoints = \{\hat{p}_i | \hat{p}_i = \diag(\initS) \initR p_i, p_i \in \priorPoints\},
    \end{split}
\vspace{-1mm}
\end{equation}
where $\diag(\cdot)$ converts a vector to a diagonal matrix.
The focalized observed point cloud $\focalizedObservedPoints$ and the focalized shape prior points $\focalizedPriorPoints$ contain full information required to predict the estimation error $(\deltaR, \deltaT, \deltaS)$. First, a PointNet-based shared encoder is used to extract features from the two focalized point clouds independently.
Then, both the extracted features are used for $\deltaR$ estimation, while the global feature of the focalized observed point cloud along with the $\initS$ are used for $\deltaT$ and $\deltaS$.
The initial estimates are updated using the predicted error $(\deltaR,\deltaT,\deltaS)$. Finally, the updated estimates are used to predict the error again in which this process is iterative and the estimations are refined progressively and continuously.

\vspace{-3mm}
\paragraph{HS-layer.} 
The Hybrid-Scope Geometric Feature Extraction Layer (HS-layer) is a simple network structure based on 3D graph convolution. It consists of two parallel paths that extract different scopes of features from the point cloud. The first path encodes the size and translation information of the target object. Meanwhile, the second path extracts outlier-robust local and global geometric features by applying graph convolution with a strategy of \textit{Receptive Fields with Feature Distance (RF-F)} metric, alongside an \textit{Outlier Robust Feature Extraction Layer (ORL)}. These properties are particularly beneficial for category-level object pose estimation tasks. For more details, please refer to~\cite{HSPose_2023_CVPR}.


\subsection{Overall Structure of \paperName}
The overall framework of our proposed object pose refinement approach is shown in Fig.~\ref{fig:framework}. This framework comprises three principal components: 1) point cloud focalization, 2) feature extraction, and 3) pose error prediction. We follow CATRE and use the same point cloud focalization module.
We apply focalization and extract features from both the observed point cloud and the shape prior by our Feature Extraction component. Then, we predict the estimation errors $(\deltaR, \deltaT, \deltaS)$ using the extracted features in the pose error prediction component.

\subsection{Graph Convolution with Learnable Affine Transformation (LAT)}
Geometric structural information is effective in estimating an object's pose for category-level object pose estimations. However, as shown in ablation study [AS-1], directly applying the 3D graph convolutions (\eg, HS-Encoder~\cite{HSPose_2023_CVPR} and 3DGCN-Encoder~\cite{3dgcn_lin_CVPR_2020}) to category-level object pose refinement tasks results in poor performance. This is due to the differences in task nature between the pose estimation and the pose refinement. In the pose estimation task, the network only needs to extract geometric structural information from a single point cloud. However, in the pose refinement framework, it requires extracting the geometric structural information from the two input point clouds as well as establishing the geometric correspondences between different object shapes. This becomes challenging due to the issue of shape variation in category pose refinement.

{To address the aforementioned problem, we propose to use learnable affine transformations (LATs). By employing LATs, the network can dynamically adjust the input point cloud and the point features which enables better establishment of geometric correspondences between the two different input shapes.}
{Specifically, we apply three LATs (as shown in the bottom left of Fig.~\ref{fig:framework}, where the Matrix Net outputs the learnable affine transformations): The first LAT $M \in \realR^3$ is applied to the input point cloud in the Euclidean space. The second LAT $\TSMatrix$ is applied to the extracted translation and size features $\TSFeature$. The third LAT $\RMatrix$ is applied to the extracted rotation feature $\RFeature$. With this approach, our method can better utilize the valuable geometric features in pose refinement.}

\subsection{Cross-Cloud Transformation (CCT) for Information Mixing}
In pose refinement, effectively blending information from the focalized observed object and the shape prior is crucial for enabling the network to align them accurately. However, in CATRE, the data from the observed point cloud and the shape prior are processed independently until the late rotation prediction stage, where they are merely concatenated, limiting the effectiveness of the alignment.
To address this problem,  we introduce a novel cross-cloud transformation mechanism that effectively mixes the geometric information from the shape of prior features into the features of the observed point cloud. In particular, we use the feature transformation matrices $\priorRMatrix$, $\priorTSMatrix$ of shape prior to transforming the features of the observed point cloud:
\vspace{-2px}
\begin{align}
    \observedRFeature &= \priorRMatrix \observedRFeature ,\\
    \observedTSFeature &= \priorTSMatrix \observedTSFeature.  
\end{align}


\subsection{Integrating Shape Prior in Pose Estimation}
The information contained in the shape prior is crucial for the network to align the observed point cloud and the shape prior. For the rotation error prediction, the information contained in the shape prior is the essential information. For the translation and size prediction, this information can also be utilized by the network to adjust the learned geometric features accordingly. Therefore, unlike CATRE, which relied solely on features extracted from the observed point cloud to predict $(\deltaT, \deltaS)$, our approach also incorporates the information from shape prior to predict them. In particular, we not only mix the information using previous CCT mechanism, but also concatenate the features from shape prior and observed point cloud to obtain mixed features in the similarly way as the rotation estimation. We utilize $\mixedTFeature$ and $\mixedSFeature$ which contain both information from shape prior and observed point cloud like $\mixedRFeature$ to predict $(\deltaT, \delta)$.


We use two pose error predictors of the same network architecture to predict the rotation error and the translation and size error, respectively. Note that the weights of these two pose error predictors are not shared. The network structure of the pose error predictor is shown in Fig.~\ref{fig:framework}. The pose predictor takes in two features, passes them through two same paths separately, and obtains two vectors in $\realR^3$.  In the translation and size branch, the pose error predictor takes in $\mixedTFeature$ and $\mixedSFeature$ and passes them through the two paths, and the output two vectors are regarded as $\deltaT$ and $\deltaS$, respectively. For the rotation error prediction, the mixed rotation features $\mixedRFeature$ are copied and passed through the two paths in the pose error predictor. The two output vectors are regarded as $r_x$ and $r_y$, where $r_x$ and $r_y$ are the first and second axes of the rotation error matrix $\deltaR$. The third column $r_z$ of $\deltaR$ can be found by:
\begin{equation}
\vspace{-2mm}
    r_z = r_x \times r_y.
\end{equation}

\begin{table*}
\begin{center}

\caption{\footnotesize \textbf{Ablation studies on REAL275.} }
\vspace{-4mm}
\caption*{\footnotesize Higher score indicates better performance. In the `Row’ column, the code in bold means the strategies taken in the final structure. In the `Method' column, the notation `$X$:$Y$' denotes module $Y$ from structure $X$, `$X$+$Y$' means add module $Y$ to $X$, and `$X \rightarrow Y$' indicates replacing $X$ with $Y$.}
\label{tbl:ablation_full}
\vspace{-3mm}
\resizebox{1\linewidth}{!}
{\footnotesize

\begin{tabular}{@{}c|l|cc|cccc|cc@{}}
\toprule

Row & Method   &$\text{IoU}_{50}$    &$\text{IoU}_{75}$     &$5^\circ 2$cm &$5^\circ 5$cm &$10^\circ 2$cm &$10^\circ 5$cm &$2$cm& $5^\circ$   \\
\midrule            
\midrule            

A0 & CATRE\cite{SPD_eccv_2020} (baseline)   &77.0 &43.6 &45.8 &54.4 &61.4 &73.1 &75.1 & 58.0\\
\midrule
\textbf{B0} & \textbf{Ours}: E0 $+$ Cross-Cloud Transformation&{\textbf{79.2}}{\scriptsize \color{mydarkgreen} $\,$2.2$\uparrow$} &{\textbf{51.8}}{\scriptsize \color{mydarkgreen} $\,$8.2$\uparrow$} & {\textbf{54.4}}{\scriptsize \color{mydarkgreen} $\,$8.6$\uparrow$} & {\textbf{60.3}}{\scriptsize \color{mydarkgreen} $\,$5.9$\uparrow$} & {\textbf{71.9}}{\scriptsize \color{mydarkgreen} $\,$10.5$\uparrow$} & {\textbf{79.4}}{\scriptsize \color{mydarkgreen} $\,$6.3$\uparrow$} & {\textbf{81.9}}{\scriptsize \color{mydarkgreen} $\,$6.8$\uparrow$} &{\textbf{64.3}}{\scriptsize \color{mydarkgreen} $\,$6.3$\uparrow$} \\
\midrule
C0 & A0: PointNet $\rightarrow$ HS-Encoder  & 71.0 & 30.1 & 41.9 & 45.9 & 60.6 & 70.3 & 71.9 & 48.7\\
C1 & A0: PointNet $\rightarrow$ 3DGCN-Encoder &- &{28.4} & {36.0} &{43.4} &{-} &{-} &68.0 &47.7 \\ 

\midrule
\textbf{D0} & A0 + prior in ST branch & {77.1} & {45.8} & {48.0} & {54.6} & {63.8} & {72.5} & {77.9} & {59.2}\\

\midrule
\textbf{E0} & D0: PointNet $\rightarrow$ HS-layer+LAT & 79.4 & 51.0 & 52.4 & 58.6 & 69.4 & 77.7 & 80.4 & 62.4\\ 
E1 & B0: No LAT on input points  & 76.1 & 39.3 & 46.6 & 53.0 & 65.4& 74.8 & 78.0 & 58.2 \\ 
E2 & B0: No LAT on features  & 78.5 & 48.8 & 47.4 & 53.0 & 67.4 & 75.0 & 80.4 & 57.4\\ 
E3 & B0: No LAT on the rotation feature  & 79.8 & 50.6 & 50.4 & 56.2 & 68.6 & 76.3 & 80.2 & 60.8 \\ 
\midrule
F0 & E0+ Global Concatenation Fusion &{77.7} &{48.4} & {47.8} & {54.5} & {67.1} & {75.2} & {80.1} &{59.4} \\

\bottomrule
\end{tabular}
}

\vspace{-12px}
\end{center}
\end{table*}
\section{Experiments}
\paragraph{Implementation details.} 
We implement and experiment with our method using an RTX 4090 GPU with a batch size of 12 and 150 training epochs. We follow CATRE~\cite{CATRE_eccv_2022} and adopt its loss functions and the basic data augmentation strategies including random dropping points, adding Gaussian noise, random pose perturbations, etc. We set the number of points for both the observed points and shape prior to be 512. We train the network using Ranger optimizer~\cite{ranger_optimizer_1_ICLA_2019, Ranger_optimizer_2_ECCV_2020, Ranger_optimizer_3_2019_NIPS} with a base learning rate of $10^{-4}$ and anneal the learning rate from the $72\%$ of the total epoch based on cosine schedule.

\vspace{-1mm}
\paragraph{Baselines.} We use CATRE as the baseline for our ablation study as it is the state-of-the-art category-level object refinement method. As CATRE did not provide the results of $\IoUFifty$, we obtained them by using their official pre-trained model and kept the rest of the reported metric scores consistent with the corresponding paper. For fair comparisons, we use the same initial estimations as CATRE, which is the pose estimation results of SPD~\cite{SPD_eccv_2020}. The result of replacing PointNet with the 3DGC-Encoder in the ablation study is provided by CATRE. We also apply our method to other state-of-the-art category-level object pose estimation approaches~\cite{HSPose_2023_CVPR, GPV-Pose_cvpr_2022, Self-DPDN_ECCV_2022} to demonstrate the effectiveness of our proposed refinement method. For the pose refinement on HS-Pose~\cite{HSPose_2023_CVPR} and RBP-Pose~\cite{RBP-Pose_ECCV_2022}, we compute the initial estimations using their official pre-trained models. The results of other methods are taken directly from their paper.

\vspace{-1mm}
\paragraph{Datasets.}
As we focus on the problem of shape variation between input object point clouds, we choose two popular category-level object pose estimation benchmarks to verify our approach, \ie, REAL275~\cite{NOCS_cvpr_2019} and CAMERA25~\cite{NOCS_cvpr_2019}. They both contain 6 object categories with multiple levels of shape complexities, \ie, bowl, can, bottle, laptop, mug, and camera. REAL275 contains 36 objects in 13 real-world scenes with 7k RGB-D images in total. Among them, 16 objects in 7 scenes are used for training, resulting in 4.3k images in training. CAMERA25 is a large synthetic RGB-D dataset. It provides 1085 objects and 275k RGB-D images for training, and 184 objects and 25k images for testing. 

\vspace{-1mm}
\paragraph{Evaluation metrics.}
Following~\cite{HSPose_2023_CVPR, CATRE_eccv_2022}, we evaluate our method using: 1) The mean average precision (mAP) of the \textit{3D Intersection over Union (IoU)} at different thresholds ($50\%$ and $75\%$) to evaluate the pose and size estimation together\footnote{Note that there was a small mistake with the IoU computation from the original benchmark evaluation code~\cite{NOCS_cvpr_2019}, we follow~\cite{CATRE_eccv_2022} to recalculate the IoU metrics for the SOTA methods.}. 2) The pose metric at $n\degree m\cm$ defines a pose as correct if the rotation error is below $n\degree$ and the translation error is below $m$ cm. Here, we use $5^\circ$, $10^\circ$, $2\text{cm}$, and $5\text{cm}$ as the thresholds.
%

\subsection{Ablation study}
To verify the proposed architecture, we conducted comprehensive ablation studies on the REAL275 dataset using the initial pose estimations from SPD~\cite{SPD_eccv_2020}. We present a quantitative comparison of our method with various key components disabled to motivate our design choices in Table~\ref{tbl:ablation_full}. Full results of the ablation study are reported in the supplementary materials.

\paragraph{[AS-1] Using geometric features directly.}
To illustrate the limitations of existing geometric-based encoder under the problem of shape variations, we replace the encoder of CATRE with two robust geometric-based point cloud convolutional structures, namely 3DGCN-Encoder~\cite{3dgcn_lin_CVPR_2020} and HS-Encoder~\cite{HSPose_2023_CVPR}. 3DGCN is a widely adopted graph convolution in existing category-level object pose estimation algorithms, while HS-Encoder is a recent architecture that achieves state-of-the-art performance in category-level object pose estimation. However, as shown in Table~\ref{tbl:ablation_full}~[C0, C1], even though both HS-Encoder and 3DGCN-Encoder are powerful in finding an object's pose from individual inputs, they failed to manage the pose refinement scenarios when there exist shape variations between the target object and the shape prior. We observed both encoders result in a performance drop when compared to the original CATRE with $\IoUSevenFive$ of $37.3\%$ (HS-Encoder) \vs $43.7\%$ (CATRE), $\fiveDfiveC$ of $43.4\%$ (3DGCN-Encoder) \vs $53.3\%$ (CATRE). 

\paragraph{[AS-2] Use prior features in translation and scale estimation.}
To validate that the information of shape prior is also important in scale and translation error prediction, we add the features of the prior shape to the scale and translation branch by using the same network architecture as the rotation branch. As shown in {Table~\ref{tbl:ablation_full}~[D0]}, incorporating shape prior information in translation and size estimation enhanced the overall performance by $2.2\%$ improvement on $\fiveDtwoC$ metric and $2.4\%$ on $\tenDtwoC$ metric.

\paragraph{[AS-3] Use learnable affine transformation (LAT) for geometric features.}
To demonstrate the effectiveness of LAT in addressing the shape variation issue, we conduct ablation studies on applying the proposed LAT to the input point cloud and the extracted geometric features in the feature space. The result is shown in Table~\ref{tbl:ablation_full}~[E0]. Compared to using geometric features directly (Table~\ref{tbl:ablation_full}~[C0]), LATs bring a significant boost on all the metrics, with $\IoUSevenFive$ improved by $\textbf{20.9\%}$, $\fiveDtwoC$ improved by $\textbf{10.5\%}$, and $\fiveDfiveC$ improved by $\textbf{12.7\%}$. The resulted network also significantly outperforms the PointNet-based encoder (Table~\ref{tbl:ablation_full}[D0]) on all the metrics with $\IoUSevenFive$ and $\tenDfiveC$ improved by $\textbf{5.2\%}$, $\fiveDtwoC$ improved by $\textbf{6.4\%}$, and $\tenDtwoC$ improved by $\textbf{5.6\%}$. These results verified the effectiveness of LAT on geometric features.

\paragraph{[AS-4] The influence of each learnable affine transformation (LAT).}
To further demonstrate the influences of each LAT, we conducted three experiments by gradually disabling LAT from the framework: 1) without the LAT on the input point cloud, 2) without applying LATs on features, and 3) without independent LATs on the rotation features, where a single LAT is used for the rotation, translation, and scale features. The results are shown in Table~\ref{tbl:ablation_full}~[E1-E3]. Compared to the directly using geometric features (Table~\ref{tbl:ablation_full}~[C0]), we show that LAT can significantly enhance the network by around $10\%$ improvements on $\IoUSevenFive$ and $7\%$ on $\fiveDfiveC$ metric in Table~\ref{tbl:ablation_full}~[E1-E3]. We verify that the combination of them consistently results in better performance.

\paragraph{[AS-5] Cross-cloud transformation (CCT) based feature fusion}
To demonstrate that it is important to have a good feature fusion strategy in the feature extraction phase, we conducted experiments on two different feature fusion strategies. One of them is the widely adopted feature fusion strategy, where the global feature of one point cloud is concatenated with the features of another point cloud and then goes through convolutional layers for feature fusion. Another one is the proposed CCT-based feature fusion. As shown in Table~\ref{tbl:ablation_full}~[F0], applying global concatenation-based fusion does not enhance the overall performance, and even results in worse performance in all the metrics with the $\fiveDtwoC$ significantly decreased by $4.6\%$, and $\fiveDfiveC$ decreased by $3.1\%$ (compared to Table~\ref{tbl:ablation_full}~[D0]). On the contrary, as shown in Table~\ref{tbl:ablation_full}~[B0], our simple CCT-based feature fusion shows its effectiveness by improving all the pose metrics by around $2.0\%$. {In Supplementary, we also visualize the influence of CCT in feature space.}
\begin{figure}[!t]
\centering
\includegraphics[width=1\linewidth, trim = 40 30 70 50 , clip]{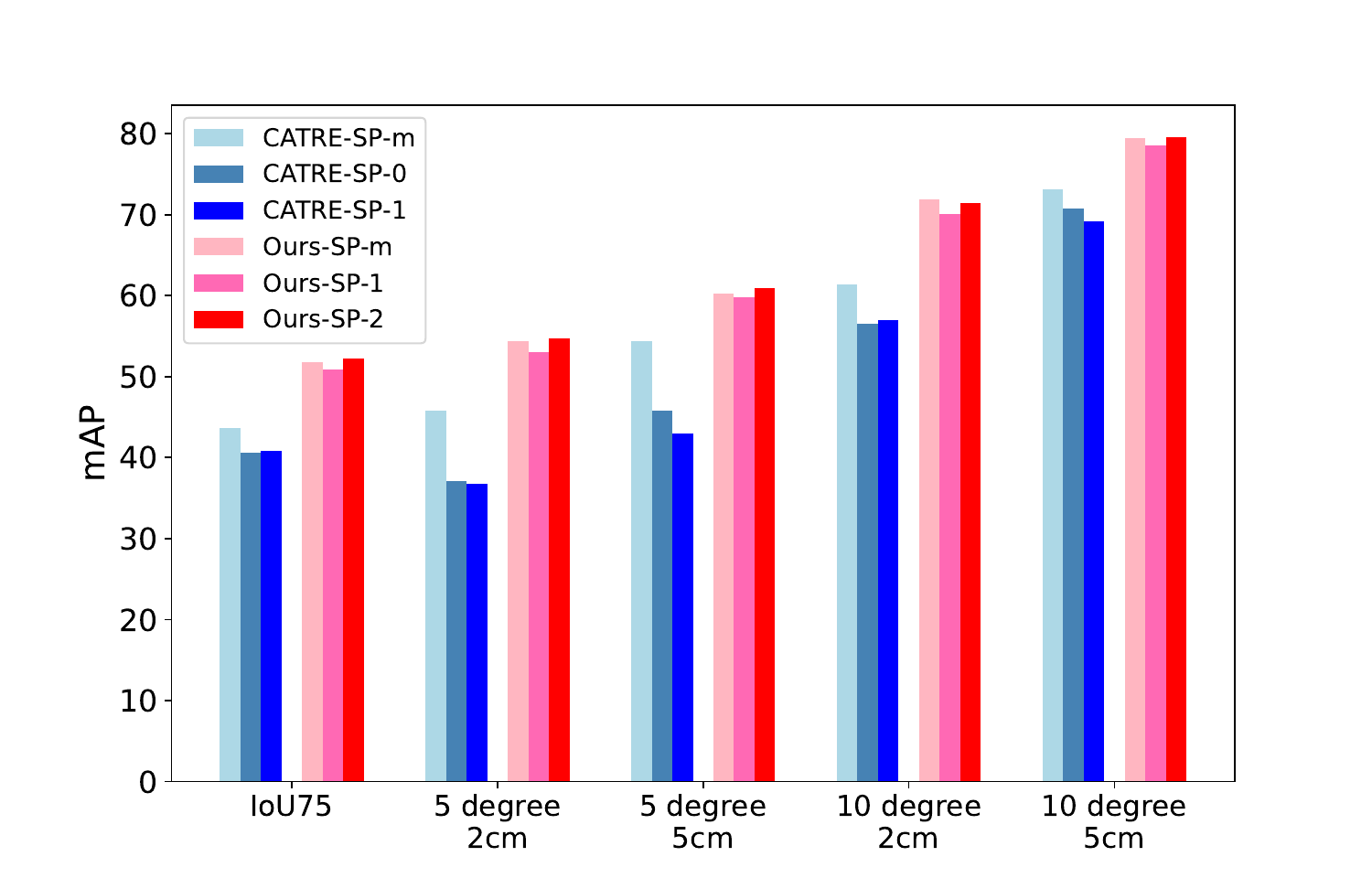}
\vspace{-17px}
\caption{\footnotesize \textbf{Performance comparison of our method and CATRE under different shape priors.}
\textit{SP-m} denotes the category's mean shape, while \textit{SP-1} and \textit{SP-2} are randomly sampled object shapes from CAMERA25.
} 
\label{fig:bar_diff_shape_prior}
\vspace{-12px}
\end{figure} 
\begin{figure*}[!t]
	\centering
	\begin{minipage}{0.06\linewidth}
	    {\footnotesize SPD}
	\end{minipage}
	\begin{minipage}{0.23\linewidth}
		\centering
		\includegraphics[width=\linewidth]{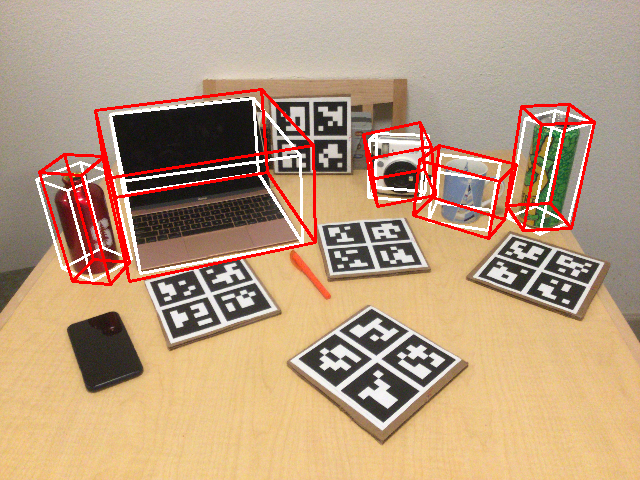}
	\end{minipage}
	\begin{minipage}{0.23\linewidth}
		\centering
		\includegraphics[width=\linewidth]{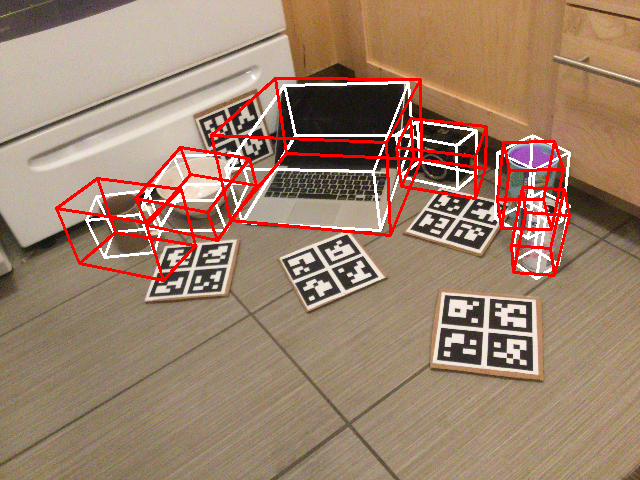}
	\end{minipage}
	\begin{minipage}{0.23\linewidth}
		\centering
		\includegraphics[width=\linewidth]{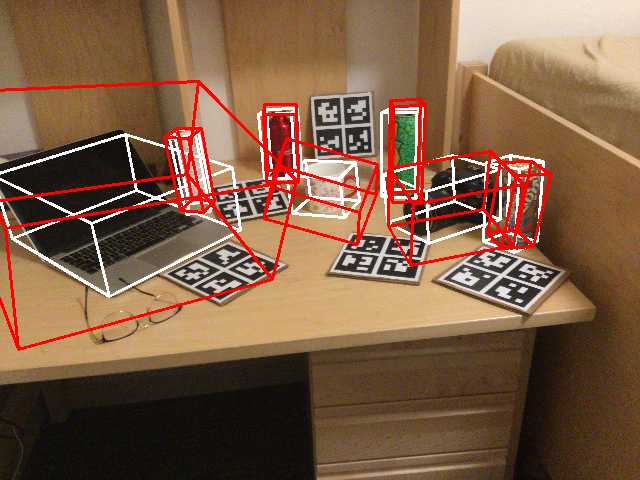}
	\end{minipage}
	\begin{minipage}{0.23\linewidth}
		\centering
		\includegraphics[width=\linewidth]{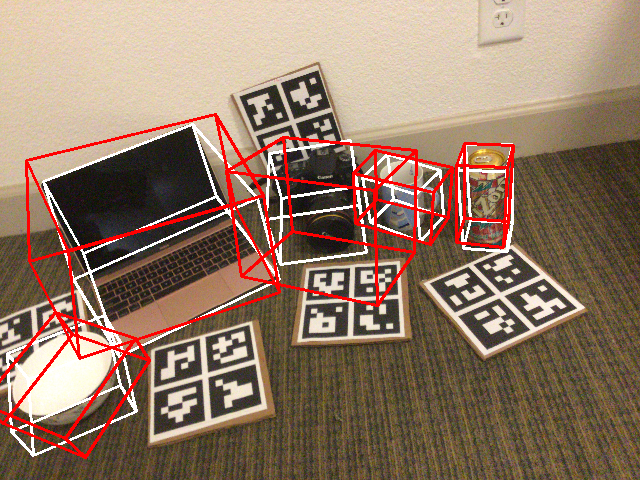}
	\end{minipage}
	
	\begin{minipage}{0.06\linewidth}
	    {\footnotesize CATRE}
	\end{minipage}
	\begin{minipage}{0.23\linewidth}
		\centering
		\includegraphics[width=\linewidth]{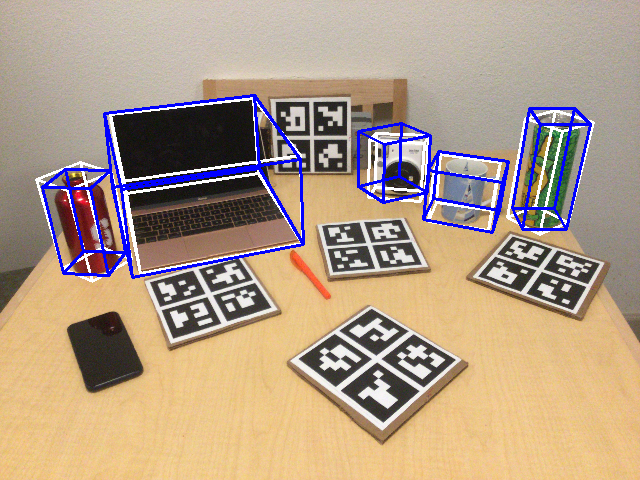}
	\end{minipage}
	\begin{minipage}{0.23\linewidth}
		\centering
		\includegraphics[width=\linewidth]{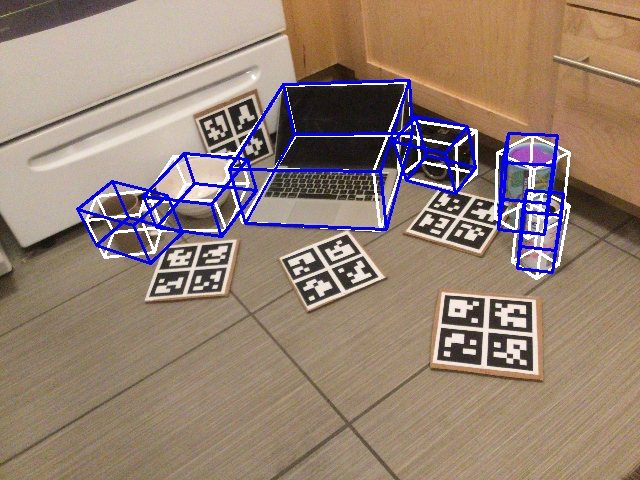}
	\end{minipage}
	\begin{minipage}{0.23\linewidth}
		\centering
		\includegraphics[width=\linewidth]{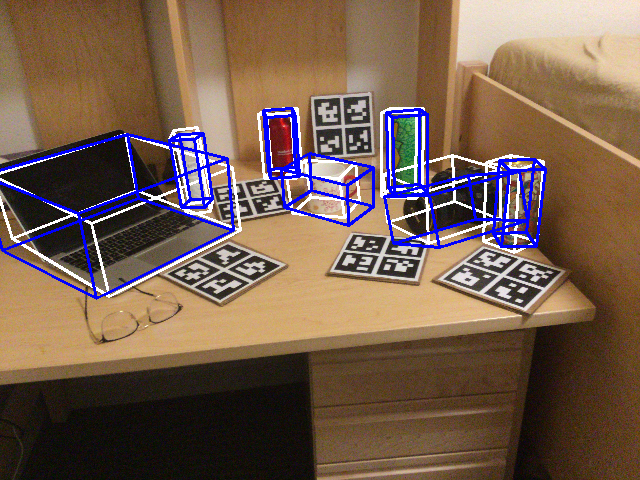}
	\end{minipage}
	\begin{minipage}{0.23\linewidth}
		\centering
		\includegraphics[width=\linewidth]{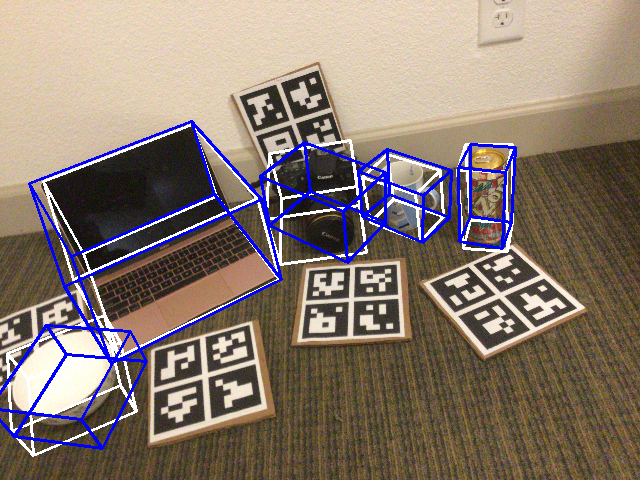}
	\end{minipage}

	\begin{minipage}{0.06\linewidth}
	    {\footnotesize \textbf{Ours}}
	\end{minipage}
	\begin{minipage}{0.23\linewidth}
		\centering
		\includegraphics[width=\linewidth]{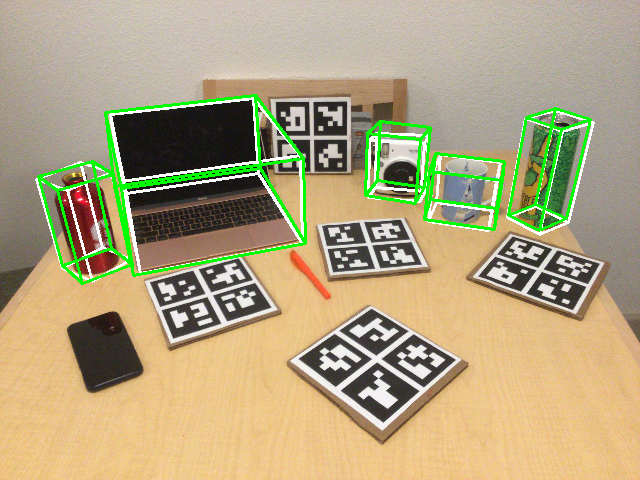}
	\end{minipage}
	\begin{minipage}{0.23\linewidth}
		\centering
		\includegraphics[width=\linewidth]{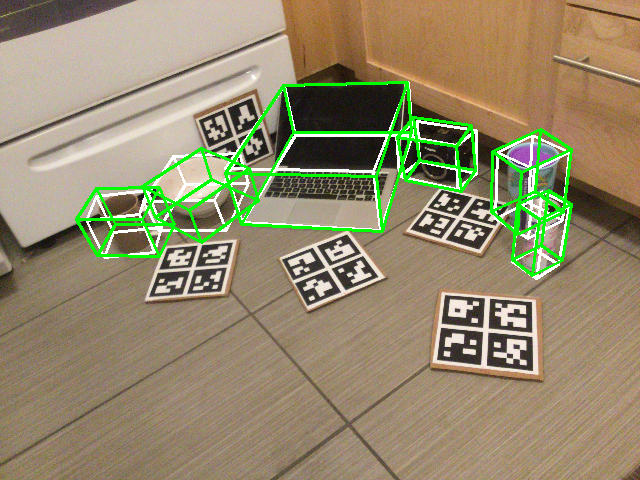}
	\end{minipage}
	\begin{minipage}{0.23\linewidth}
		\centering
		\includegraphics[width=\linewidth]{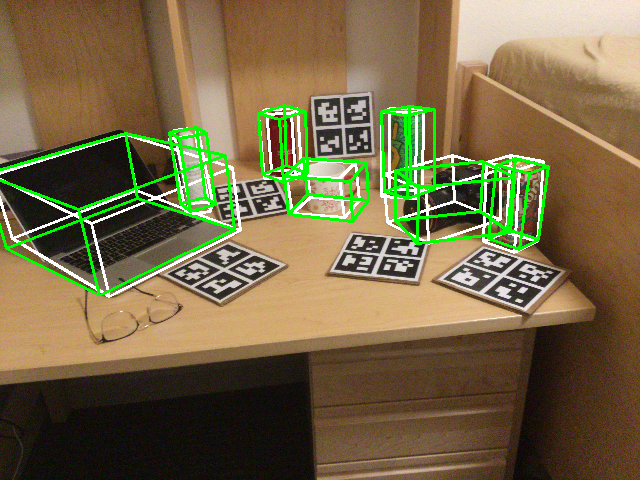}
	\end{minipage}
	\begin{minipage}{0.23\linewidth}
		\centering
		\includegraphics[width=\linewidth]{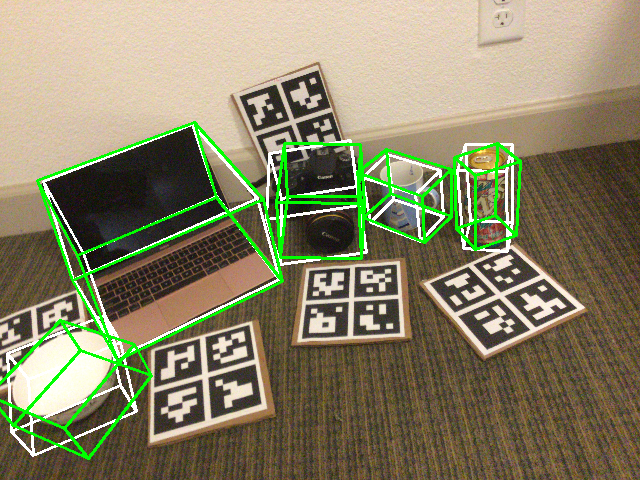}
	\end{minipage}

        \vspace{-2mm}
	\caption{\footnotesize \textbf{Qualitative comparison of proposed (row \#3) and baseline (row \#2) methods using SPD (row \#1) as initial estimation.} Ground truth shown with white lines. Note that the estimated rotations of symmetric objects (\eg bowl, bottle, and can) are considered correct if the symmetry axis is aligned.}
\label{fig:qualitative_main}
\vspace{-10px}
\end{figure*}

\paragraph{[AS-6] Handle shape variations.}
To demonstrate that the proposed method can handle the shape variations, we replaced the original shape prior with two randomly sampled models from CAMERA25 training set and trained on REAL275. Note that the original shape prior represents the mean shape of a category, and the new models are randomly sampled. Therefore, there are larger shape variations with certain target objects. As shown in Fig.~\ref{fig:bar_diff_shape_prior}, CATRE performs best when using the mean shape of the category, while its performance drops dramatically on the randomly sampled shape priors. In contrast, our method exhibits robustness to shape variations introduced by different shape priors and consistently delivers strong performance.

\paragraph{[AS-7] Refinement with different initial estimations.}
To demonstrate our model robustness to different initial estimations, we compare the proposed method with CATRE on different initial estimations generated by category-level object pose estimation methods with varying performance, including HS-Pose~\cite{HSPose_2023_CVPR}, GPVPose~\cite{GPV-Pose_cvpr_2022}, and the Self-DPDN~\cite{Self-DPDN_ECCV_2022}. 
As shown in Table~\ref{tbl:refinement_comparison}, our method consistently improves the initial estimations by a large margin across different metrics. For example, we improved the SelfDPND and GPV-Pose on strict $\fiveDtwoC$ metric by 9.6\% and 15.4\%, respectively. We boosted the $\IoUSevenFive$ of the GPV-Pose and the HS-Pose by 26.5\% and 15.2\%, respectively. However, the baseline CATRE fails to refine the initial poses of Self-DPDN. Also, CATRE reaches its limit when refining high-accuracy initial estimations such as the initial poses generated by HS-Pose, resulting in performance drops on $\tenDfiveC$ and $\tenDtwoC$. The comparison results demonstrated the robustness and capability of our method to different initial estimations. We report full details in the supplementary.

\subsection{Generalizabily test on the CAMERA25 dataset}
In real-world applications, category-level algorithms often need to generalize across diverse testing scenarios, encountering a larger number of objects than represented in their training sets. 
To simulate this problem setting, we choose CAMERA25 as it provides more than 25K RGB-D testing images. We train our model using only a mini set~(2\% and 4\%) of the CAMERA25 training data, resulting in a training set of around 5K and 10K images from a total of 275K images. As shown in Table~\ref{tbl:generalizability_test_on_camera25}, the performance of the baseline method decreased dramatically when using 2\% of the training images, with $\IoUSevenFive$ dropped by \textbf{12.9\%} and $\fiveDtwoC$ dropped by \textbf{9.0\%}~(see~[B0]). On the contrary, our method exhibits a much higher performance when using small datasets for training. Specifically, our method can already outperforms the fully-trained CATRE with only \textbf{2\%} of images~(see~[B1]). We also report the performance using 4\% images of the training set in Table~\ref{tbl:generalizability_test_on_camera25}. It can be seen that, despite CATRE use full CAMERA25 training set, our method outperforms it with $\IoUSevenFive$ and $\fiveDtwoC$ improved by 3.1\% and 3.6\%, respectively.

\begin{table}[!t]
\centering
\caption{\footnotesize \textbf{Comparison of the pose refinement with the baseline method CATRE on REAL275 with different initial estimations.} Each comparison group contains 3 methods: the initial pose estimation method, refinement using CATRE, and refinement using our method, respectively. }
\vspace{-7px}
\label{tbl:refinement_comparison}
\begin{adjustbox}{max width=\linewidth}
\begin{tabular}{l|c|cccc}
    \toprule 
    Method &  $\text{IoU}_{75}$ & $5^\circ 2$cm & $5^\circ 5$cm & $10^\circ 2$cm & $10^\circ 5$cm \\
    \midrule
    \midrule
    Self-DPDN \cite{Self-DPDN_ECCV_2022} & 42.2 & 44.3 & 50.9 & 65.1 & 78.6\\
    Self-DPDN + CATRE & 0.0{\footnotesize \color{mydarkred} $\,$42.2$\downarrow$} & 0.3{\footnotesize \color{mydarkred} $\,$44.0$\downarrow$} & 5.1{\footnotesize \color{mydarkred} $\,$45.8$\downarrow$} & 0.4{\footnotesize \color{mydarkred} $\,$64.7$\downarrow$} & 6.9{\footnotesize \color{mydarkred} $\,$71.7$\downarrow$} \\
    Self-DPDN + Ours & 49.7{\footnotesize \color{mydarkgreen} $\,$7.5$\uparrow$} & 53.9{\footnotesize \color{mydarkgreen} $\,$9.6$\uparrow$} & 60.1{\footnotesize \color{mydarkgreen} $\,$9.2$\uparrow$} & 75.0{\footnotesize \color{mydarkgreen} $\,$9.9$\uparrow$} & 82.8{\footnotesize \color{mydarkgreen} $\,$4.2$\uparrow$} \\
    \midrule
    GPV-Pose \cite{GPV-Pose_cvpr_2022} & 23.1* & 32.0 & 42.9 & 55.0 & 73.3\\
    GPV-Pose + CATRE & 42.6{\footnotesize \color{mydarkgreen} $\,$19.5$\uparrow$} & 39.7{\footnotesize \color{mydarkgreen} $\,$7.7$\uparrow$} & 54.1{\footnotesize \color{mydarkgreen} $\,$11.2$\uparrow$} & 57.1{\footnotesize \color{mydarkgreen} $\,$2.1$\uparrow$} & 78.0{\footnotesize \color{mydarkgreen} $\,$4.7$\uparrow$}\\
    GPV-Pose + Ours & 49.6{\footnotesize \color{mydarkgreen} $\,$26.5$\uparrow$} & 47.4{\footnotesize \color{mydarkgreen} $\,$15.4$\uparrow$} & 57.8{\footnotesize \color{mydarkgreen} $\,$14.9$\uparrow$} & 68.1{\footnotesize \color{mydarkgreen} $\,$13.1$\uparrow$} & 81.2{\footnotesize \color{mydarkgreen} $\,$7.9$\uparrow$}\\
    \midrule
    HS-Pose \cite{HSPose_2023_CVPR} & 39.1* & 46.5 & 55.2 & 68.6 & 82.7\\
    HS-Pose + CATRE & 47.1{\footnotesize \color{mydarkgreen} $\,$8.0$\uparrow$} & 48.7{\footnotesize \color{mydarkgreen} $\,$2.2$\uparrow$} & 59.1{\footnotesize \color{mydarkgreen} $\,$3.9$\uparrow$} & 67.8{\footnotesize \color{mydarkred} $\,$0.8$\downarrow$} & 81.2{\footnotesize \color{mydarkred} $\,$1.5$\downarrow$} \\
    HS-Pose + Ours & 54.3{\footnotesize \color{mydarkgreen} $\,$15.2$\uparrow$} & 51.7{\footnotesize \color{mydarkgreen} $\,$5.2$\uparrow$} & 59.6{\footnotesize \color{mydarkgreen} $\,$4.4$\uparrow$} & 74.3{\footnotesize \color{mydarkgreen} $\,$5.7$\uparrow$} & 83.8{\footnotesize \color{mydarkgreen} $\,$1.1$\uparrow$}\\
    \bottomrule
\end{tabular}
\end{adjustbox}
\vspace{-10px}
\end{table}

\begin{table*}[htbp]
\begin{center}
\caption{\footnotesize \textbf{The generalizability test on the CAMERA25 dataset.} }
\label{tbl:generalizability_test_on_camera25}
\vspace{-2mm}
\resizebox{0.7\linewidth}{!}
{\footnotesize

\begin{tabular}{@{}c|l|c|c|cccc|cc@{}}
\toprule

Row & Method &Train Data Size&$\text{IoU}_{75}$ &$5^\circ 2$cm &$5^\circ 5$cm &$10^\circ 2$cm &$10^\circ 5$cm &$2$cm& $5^\circ$   \\
\midrule            
\midrule            
A0 & CATRE &275K &{76.1} & \underline{75.4} &{80.3} &83.3 &89.3 & - & - \\ 
\midrule       
B0 & CATRE &5K &{63.2} & {66.4} &{72.3} &79.4 &87.4 &88.8 & 73.3 \\ 
B1 & \textbf{Ours} &5K   &\underline{77.5} & \underline{75.4} &\underline{81.1} &\underline{83.4} &\underline{90.0} &\underline{91.0} & \underline{82.3} \\ 
\midrule       
C0 & CATRE &10K &{66.5} & {69.7} &{75.5} &{81.8} &{89.1} & 89.9 & 76.7 \\ 
C1 & \textbf{Ours} &10K  &{\textbf{79.2}} & \textbf{{77.9}} &\textbf{{84.0}} &\textbf{{83.8}} &\textbf{{90.5}} &\textbf{92.0} & \textbf{85.4 }\\ 

\bottomrule
\end{tabular}
}
\vspace{-10px}

\end{center}
\end{table*}

\begin{table}[htbp]

\centering

\caption{\footnotesize \centering\textbf{Comparison with other methods on REAL275.}}
\vspace{-7px}
\label{tbl:sota_perform_real275}
\resizebox{1\linewidth}{!}
{\begin{tabular}{@{}l|c|cccc@{}}
\toprule
Method & $\text{IoU}_{75}$ & $5^\circ 2$cm & $5^\circ 5$cm & $10^\circ 2$cm & $10^\circ 5$cm  \\ 

\midrule
\midrule

NOCS \cite{NOCS_cvpr_2019}  &  9.4 & 7.2 & 10.0 & 13.8 & 25.2 \\
DualPoseNet \cite{DualPoseNet_iccv_2021} & 30.8 & 29.3 & 35.9 & 50.0 & 66.8 \\
CR-Net \cite{CR-Net_iros_2021} &  33.2 & 27.8 & 34.3 & 47.2 & 60.8 \\
SGPA \cite{SGPA_iccv_2021} &  37.1 & 35.9 & 39.6 & 61.3 & 70.7 \\
RBP-Pose \cite{RBP-Pose_ECCV_2022}  & 24.5 & 38.2 & 48.1 & 63.1 & 79.2 \\
GPV-Pose \cite{GPV-Pose_cvpr_2022}  & 23.1 & 32.0 & 42.9 & 55.0 & 73.3 \\
HS-Pose \cite{HSPose_2023_CVPR}  & {39.1} & \underline{46.5} & \underline{55.2} & \underline{68.6} & \textbf{82.7}\\ 
SPD$^*$ \cite{SPD_eccv_2020}   & 27.0 & 19.1 & 21.2 & 43.5 & 54.0 \\ 
\midrule
SPD$^*$$+$CATRE \cite{CATRE_eccv_2022} &  \underline{43.6} & 45.8 & 54.4 & 61.4 & 73.1 \\
SPD$^*$$+$\textbf{Ours} & \textbf{ 51.8} &\textbf{ 54.4} & \textbf{60.3} & \textbf{71.9} &  \underline{79.4} \\ 
\bottomrule
\end{tabular}}
\vspace{-10px}
\end{table}

\subsection{Comparison with state-of-the-arts}
\paragraph{REAL275.} We conduct pose refinement on SPD~\cite{SPD_eccv_2020} using the proposed approach and compare the resulting performance with the state-of-the-art category-level object pose estimation and refinement methods. As shown in Table~\ref{tbl:sota_perform_real275}, our method significantly improves the performance of SPD on all the metrics, with $\fiveDfiveC$ enhanced by \textbf{39.1\%}, $\fiveDfiveC$ improved by \textbf{35.3\%}, $\IoUSevenFive$ enhanced by \textbf{24.8\%}, $\tenDtwoC$ improved by \textbf{24.4\%}, and $\tenDfiveC$ improve by \textbf{25.4\%}. 
In comparison to our baseline, CATRE, our proposed method demonstrates a substantial improvement across various performance metrics. Specifically, we observe a remarkable enhancement of \textbf{10.5\%} in $\tenDtwoC$, \textbf{8.6\%} in $\fiveDtwoC$, \textbf{7.2\%} in $\IoUSevenFive$, and \textbf{6.3\%} in $\tenDfiveC$. 
In addition, compared with SOTA pose estimation methods, the pose estimation results of applying our proposed refinement method on SPD significantly outperformed these methods's results by a large margin. Specifically, the estimation results of applying our method on SPD rank top on 4 out of 5 metrics and rank second on the rest metric, and achieved a \textbf{7.9\%} increase on the $\fiveDtwoC$ metric and \textbf{5.1\%} enhancement on $\fiveDfiveC$. 
It is worth noting that the purpose of this section is not to compare refinement and estimation methods, as they are designed to address different problems. Instead, our objective here is to demonstrate how our proposed refinement approach can improve the performance of existing pose estimation methods.
Therefore, even though our refinement on HS-Pose produced better performance, as mentioned in the ablation study, we choose to refine weaker initial estimations to show the capability of our approach.

\vspace{-10px}
\paragraph{CAMERA25.} Our method outperforms the state-of-the-art methods using only 2\% of the training image set. For more details, please refer to the supplementary.

\begin{figure}[!t]
	\centering
	\begin{minipage}{0.12\linewidth}
	    {\footnotesize CATRE}
	\end{minipage}
	\begin{minipage}{0.16\linewidth}
		\centering
		\includegraphics[width=\linewidth]{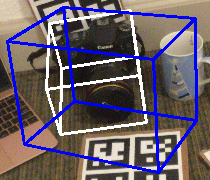}
	\end{minipage}
	\begin{minipage}{0.16\linewidth}
		\centering
		\includegraphics[width=\linewidth]{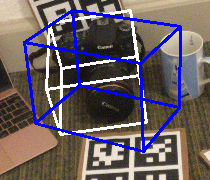}
	\end{minipage}
 	\begin{minipage}{0.16\linewidth}
		\centering
		\includegraphics[width=\linewidth]{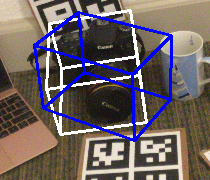}
	\end{minipage}
	\begin{minipage}{0.16\linewidth}
		\centering
		\includegraphics[width=\linewidth]{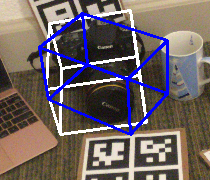}
	\end{minipage}
 	\begin{minipage}{0.16\linewidth}
		\centering
		\includegraphics[width=\linewidth]{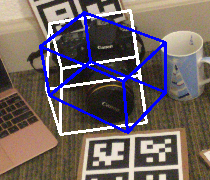}
	\end{minipage}

	\begin{minipage}{0.12\linewidth}
	    {\footnotesize \textbf{Ours}}
	\end{minipage}
	\begin{minipage}{0.16\linewidth}
		\centering
		\includegraphics[width=\linewidth]{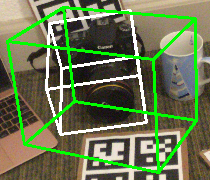}
	\end{minipage}
	\begin{minipage}{0.16\linewidth}
		\centering
		\includegraphics[width=\linewidth]{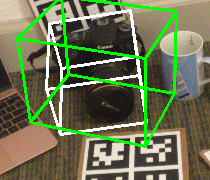}
	\end{minipage}
 	\begin{minipage}{0.16\linewidth}
		\centering
		\includegraphics[width=\linewidth]{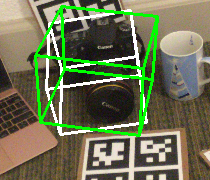}
	\end{minipage}
	\begin{minipage}{0.16\linewidth}
		\centering
		\includegraphics[width=\linewidth]{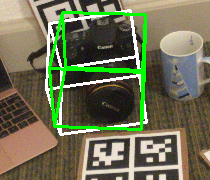}
	\end{minipage}
 	\begin{minipage}{0.16\linewidth}
		\centering
		\includegraphics[width=\linewidth]{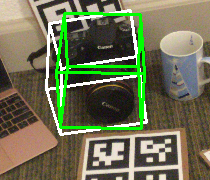}
	\end{minipage}

	\begin{minipage}{0.12\linewidth}
	    {\footnotesize {\color{white} aaaa}}
	\end{minipage}
	\begin{minipage}{0.16\linewidth}
		\centering
            {\footnotesize Initial Pose}
	\end{minipage}
	\begin{minipage}{0.16\linewidth}
		\centering
            {\footnotesize Iter 1}
	\end{minipage}
 	\begin{minipage}{0.16\linewidth}
		\centering
            {\footnotesize Iter 2}
	\end{minipage}
	\begin{minipage}{0.16\linewidth}
		\centering
            {\footnotesize Iter 3}
	\end{minipage}
 	\begin{minipage}{0.16\linewidth}
		\centering
            {\footnotesize Iter 4}
	\end{minipage}

    \vspace{-2mm}
	\caption{\footnotesize \textbf{Comparison of proposed (row \#2) and baseline (row \#1) methods) during a complete refinement iteration, both using SPD as initial estimation.} The ground truth is represented by white lines.}
\label{fig:qualititive_iter}
\vspace{-4mm}
\end{figure}

\vspace{-10px}
\paragraph{Qualitative examples.} 
We provide qualitative comparisons of the pose estimation results by SPD, CATRE on SPD, and our method on SPD in Fig.~\ref{fig:qualitative_main} and~\ref{fig:qualititive_iter}. As shown in Fig.~\ref{fig:qualitative_main}, our method on SPD achieves the best size and pose estimations. 
In particular, by considering the first column of Fig.~\ref{fig:qualitative_main}, all of the comparison methods struggle to estimate the orientation of the camera category.
We also provide qualitative examples on the iteration process in Fig.~\ref{fig:qualititive_iter}. 
Our method demonstrates faster convergence and more accurate final result than CATRE.



\vspace{-5px}
\section{Conclusion}
\noindent{In this work, we proposed a novel category-level object pose refinement method which targeted at addressing the challenge of shape variation. We shown that the geometric structural information can be aligned by our adaptive affine transformations. We also demonstrated that the cross-cloud transformation mechanism can efficiently merges information from distinct point clouds. We further incorporated shape prior information and observed improvements in translation and size predictions. We verified that each of our technical components contributed meaningfully through extensive ablations. We believe our method sets a strong baseline for future study and opens up new possibilities to handling more complex shapes, \ie articulated objects.
}

\section*{Acknowledgments}
This research was supported in part by the MSIT (Ministry of Science and ICT), Korea, under the ITRC (Information Technology Research Center) support program (IITP-2024-2020-0-01789) supervised by the IITP (Institute for Information \& Communications Technology Planning \& Evaluation), National Natural Science Foundation of China under Grant No.62073159, and the Shenzhen Key Laboratory of Control Theory and Intelligent Systems, under grant No.ZDSYS20220330161800001. 
{
    \small
    \bibliographystyle{ieeenat_fullname}
    \bibliography{ref}
}


\end{document}


\maketitle
\section{About the {Runtime}}
On a machine with an Intel 13900k CPU and a Nvidia RTX 4090 GPU, the speed of our proposed method is 67.5 FPS for 1 iteration, and 22.3 FPS when using 4 iterations. 

\section{{Effect of number of iterations}}
We find that the performance of our proposed method saturates after 4 iterations. Therefore, we set the iteration number to 4 for our experiments. We provide a line graph to show the performance changes of our method and CATRE~\cite{CATRE_eccv_2022} during the iteration in Fig.~\ref{fig:line_iter}
{We show that our proposed method consistently outperforms the baseline method and saturates after 4 iterations in both figures.}

\section{Ablation Studies}
\begin{figure}[!t]
	\centering
	\begin{minipage}{0.49\linewidth}
		\centering
		\includegraphics[width=\linewidth, trim = 0 0 40 80, clip]{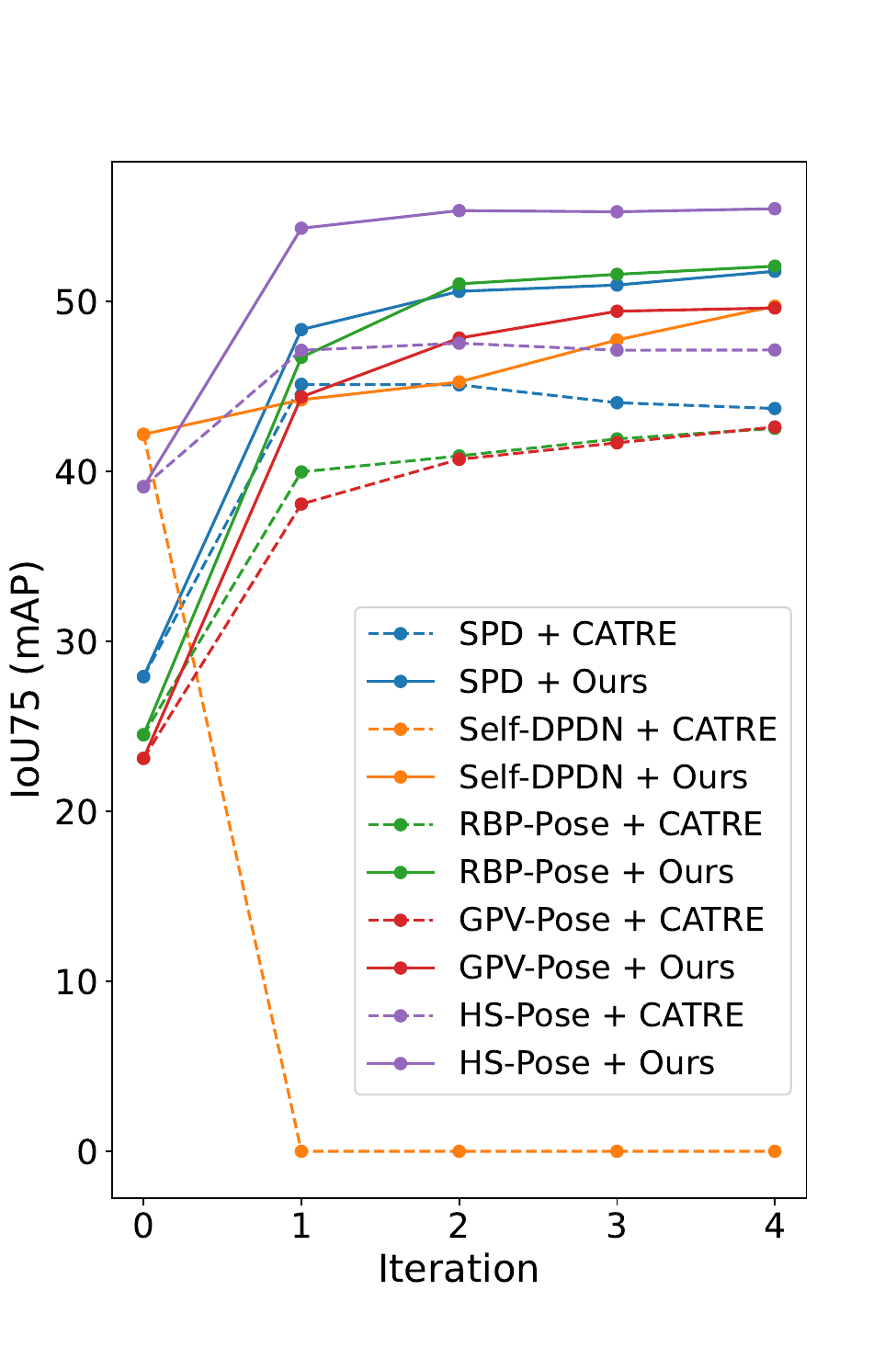}
	\end{minipage}
	\begin{minipage}{0.49\linewidth}
		\centering
		\includegraphics[width=\linewidth, trim = 0 0 40 80, clip]{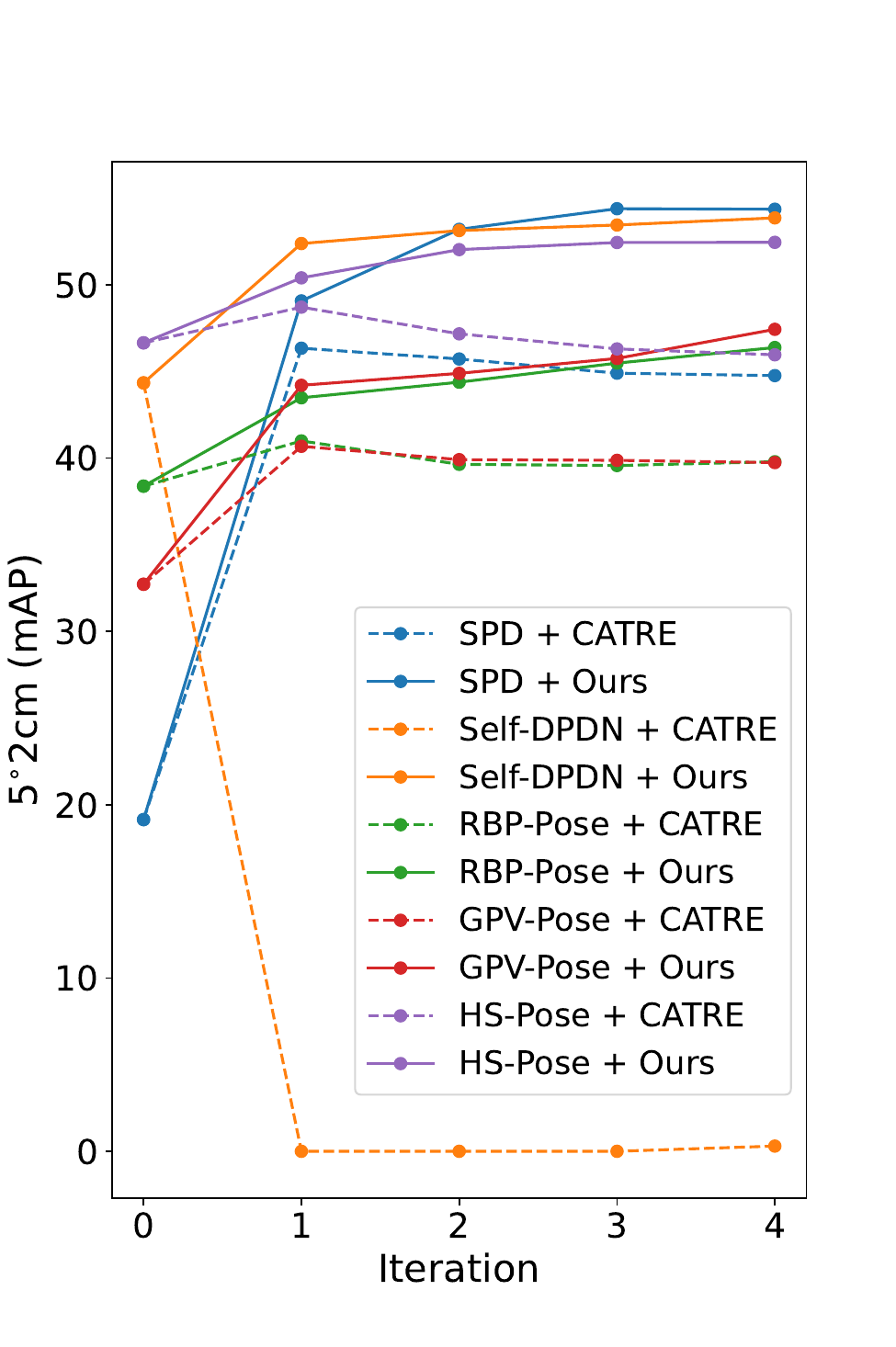}
	\end{minipage}

        \vspace{-2mm}
 	\begin{minipage}{0.49\linewidth}
		\centering
            {\small (a)}
	\end{minipage}
	\begin{minipage}{0.49\linewidth}
		\centering
            {\small (b)}
	\end{minipage}

        \vspace{-2mm}
	\caption{\footnotesize \textbf{Comparison between CATRE and our method on different initial estimations across different refining iterations.} (a) $\IoUSevenFive$ performance comparison. (b) $\fiveDtwoC$ performance comparison. Our methods are shown in solid lines and CATRE's are in dashed lines. Iteration 0 shows the performance of the initial estimations.}
\label{fig:line_iter}
\vspace{-2mm}
\end{figure}

\paragraph{Refinement with different initial estimations.}
Apart from the table provided in the main paper, we visually show the robustness of our method on different initial estimations generated from 5 pose estimation methods~\cite{SPD_eccv_2020, Self-DPDN_ECCV_2022, RBP-Pose_ECCV_2022, GPV-Pose_cvpr_2022, HSPose_2023_CVPR} with ranging performance. As shown in Fig.~\ref{fig:line_iter}, 
our method keeps improving the performance of the initial estimations, while CATRE~\cite{CATRE_eccv_2022} failed when refining the initial estimations from Self-DPDN~\cite{Self-DPDN_ECCV_2022}. Additionally, our method keeps improving during the iterations, while CATRE's performance starts to decrease after one iteration (see the dashed lines in Fig.~\ref{fig:line_iter}).

\paragraph{The effect of CCT.}
\begin{figure}[tp]
\centering
\includegraphics[width=1\linewidth, trim = 00 20 00 41, clip]{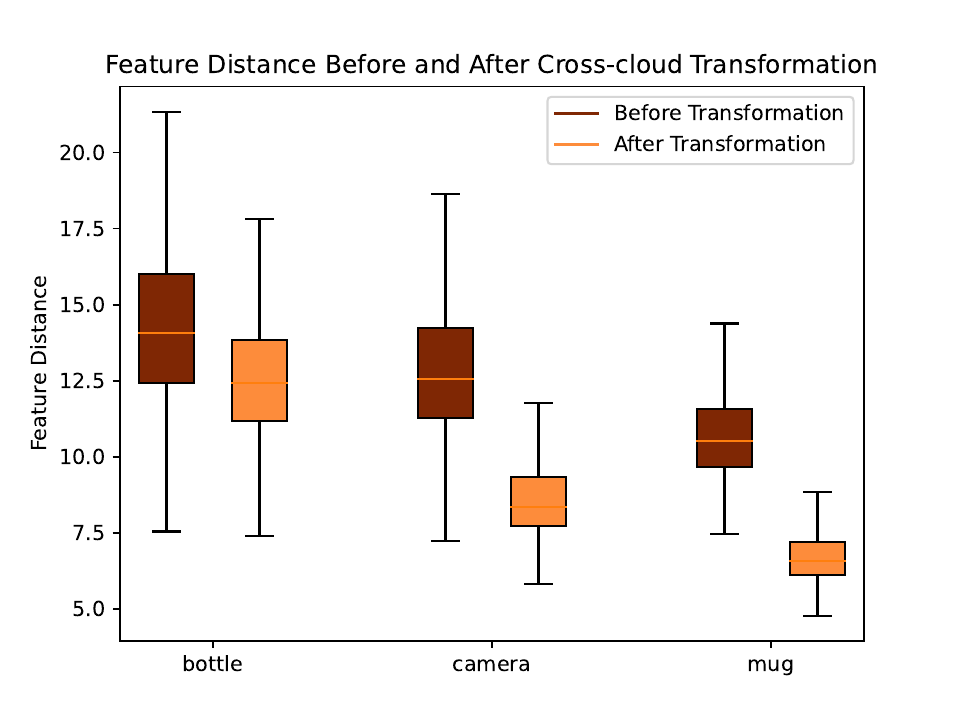}
\vspace{-18px}
\caption{\footnotesize \textbf{Feature distances between the shape prior and the input point cloud before and after applying the cross-cloud transformation.} } 
\label{fig:cct_feature_distance}
\end{figure} 

To demonstrate the effect of CCT, we show a statistics plot of feature distances before and after CCT on objects with different shape complexities of the CAMERA25 test set. In this experiment, the initial pose of the shape prior is aligned with the ground truth pose to guarantee that the observed variations in feature distance are solely attributable to differences in shape. As shown in Fig~\ref{fig:cct_feature_distance}, the feature distance between the shape prior and the input target shrinks significantly after applying CCT.



\section{Generalizability test on CAMERA25}
\label{sec:mini_dataset_camera25}
\paragraph{More Results.}
To test the generalizability of our method when trained on a small dataset and tested on a large dataset, we randomly sample datasets from the CAMERA25 training dataset at different ratios (2\%, 4\%, and 6\%). This yields training sizes of 5k, 10k, 15k. We show the results of the generalizability test in Table~\ref{tbl:mini_dataset_camera25}. We observe that our method, trained only using 2\% of the train set, can already outperform a fully trained CATRE on all training data. Also, our performance becomes stable when using 4\% of the train set (see Table~\ref{tbl:mini_dataset_camera25}~[C1, D1]), while CATRE requires additional training data for better performance. Since our performance became stable, we did not test on larger data sizes.

\paragraph{Experiment settings:} To ensure the distribution of different categories in the sampled mini datasets, we control the image number of each object in the sampled datasets: 1) 5 images per object for the 2\% train set, 2) 10 images for the 4\% train set, and 3) 15 images for the 6\% train set.

\begin{table}[htbp]
\begin{center}

\caption{\textbf{The generalizability test on the CAMERA25 dataset.} }
\vspace{-4mm}
\caption*{\footnotesize Higher score means better performance. Overall best results are in bold, and the second-best results are underlined. The training data size is denoted as \textit{T. Size}.}
\label{tbl:mini_dataset_camera25}
\vspace{-3mm}
\resizebox{1\linewidth}{!}
{\footnotesize

\begin{tabular}{@{}c|l|c|c|cccc@{}}
\toprule

Row & Method &T. Size&$\text{IoU}_{75}$ &$5^\circ 2$cm &$5^\circ 5$cm &$10^\circ 2$cm &$10^\circ 5$cm \\
\midrule            
\midrule            
A0 & CATRE &275k &{76.1} & {75.4} &{80.3} &83.3 &89.3\\ 
\midrule       
B0 & CATRE &5k &{63.2} & {66.4} &{72.3} &79.4 &87.4 \\ 
B1 & Ours &5k  &{77.5} & {75.4} &{81.1} &{83.4} &\underline{90.0} \\ 
\midrule       
C0 & CATRE &10k&{66.5} & {69.7} &{75.5} &{81.8} &{89.1} \\ 
C1 & Ours &10k &{\textbf{79.2}} & \underline{{77.9}} &\underline{{84.0}} &\textbf{{83.8}} &\textbf{{90.5}} \\ 
\midrule       
{D0} & CATRE &15k  &{69.7} & {73.2} &{78.8} &{82.6} &{89.4} \\ 
D1 & Ours &15k &\underline{78.1} & \textbf{78.0} &\textbf{84.1} &\underline{83.6} &\textbf{90.5} \\ 

\bottomrule
\end{tabular}
}
\vspace{-10px}
\end{center}
\end{table}

\section{Detailed Network Architectures}
\begin{figure}[htbp]
\centering
\includegraphics[width=1\linewidth]{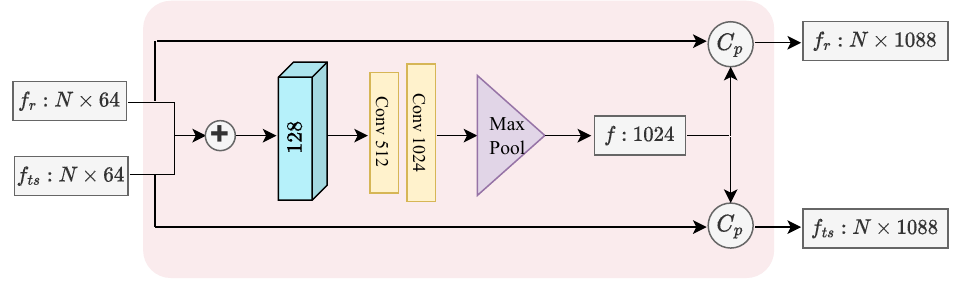}
\vspace{-6mm}
\caption{\footnotesize \textbf{Structure of the global extractor.}} 
\label{fig:global_extractor}
\end{figure}

The network structure of the HS Feature Extractor and the Pose Error Predictor is shown in Fig.~2 of the main paper. The structure of the Pose Error Predictor for $\deltaR$ estimation and the $\deltaT, \deltaS$ estimation are identical, we follow the CATRE~\cite{CATRE_eccv_2022} and use 3 Convolution-1D layers with permutation before the final layer to generate the pose errors. For the Matrix Net, we follow PointNet~\cite{Pointnet_2017_CVPR} first use 3 Convolution-1D layers with $[64, 128, 1024]$ output dimensions and a kernel size of 1 to extract the dense point features, then the features going through a maximum pooling layer and 3 liner layers with [512, 256, $\LATFeature$] to generate the matrix. For the first Matrix Net that generates the adaptive affine transformation (LAT) for the input point cloud, $\LATFeature$ is 9. For the second Matrix Net, $\LATFeature$ is 8192, as it outputs two LATs with the matrix size of $\realR^{64\times 64}$. In the final structure of the \paperName, we use two HS-layers to replace the first two Convolution-1D layers in the second Matrix Net, which in our experiments, show slightly better results than without HS-layers (See Table~\ref{tbl:sup_ablation}~[B0, G0] for the performance comparison).
The structure of the Global Feature Extractor is shown in Fig.~\ref{fig:global_extractor}, we use 1 layer of HS-layer and 2 Convolution-1D layers with the output size of $[128, 512, 1024]$ to extract dense point features, and then apply maximum pooling to get the global feature. Finally, the global feature is concatenated with the input features for the outputs.
\begin{table*}
\begin{center}

\caption{\textbf{Ablation studies on REAL275.} }
\vspace{-4mm}
\caption*{\footnotesize Higher score means better performance. Overall best results are in bold. Row's code in bold means the strategies taken in the final structure.}
\label{tbl:sup_ablation}
\vspace{-3mm}
\resizebox{1\linewidth}{!}
{\footnotesize

\begin{tabular}{@{}c|l|cc|cccc|cc@{}}
\toprule

Row & Method   &$\text{IoU}_{50}$    &$\text{IoU}_{75}$     &$5^\circ 2$cm &$5^\circ 5$cm &$10^\circ 2$cm &$10^\circ 5$cm &$2$cm& $5^\circ$   \\
\midrule            
\midrule            

A0 & CATRE~\cite{CATRE_eccv_2022} (baseline)   &77.0 &43.6 &45.8 &54.4 &61.4 &73.1 &75.1 & 58.0\\
\midrule
\textbf{B0} & \textbf{Ours}: E0+Cross-Cloud Transformation&{79.2}{\scriptsize \color{mydarkgreen} $\,$\textbf{2.2}$\uparrow$} &{\textbf{51.8}}{\scriptsize \color{mydarkgreen} $\,$\textbf{8.2}$\uparrow$} & {\textbf{54.4}}{\scriptsize \color{mydarkgreen} $\,$\textbf{8.6}$\uparrow$} & {\textbf{60.3}}{\scriptsize \color{mydarkgreen} $\,$\textbf{5.9}$\uparrow$} & {\textbf{71.9}}{\scriptsize \color{mydarkgreen} $\,$\textbf{10.5}$\uparrow$} & {\textbf{79.4}}{\scriptsize \color{mydarkgreen} $\,$\textbf{6.3}$\uparrow$} & {\textbf{81.9}}{\scriptsize \color{mydarkgreen} $\,$\textbf{6.8}$\uparrow$} &{\textbf{64.3}}{\scriptsize \color{mydarkgreen} $\,$\textbf{6.3}$\uparrow$} \\
\midrule
C0 & A0: PointNet $\rightarrow$ HS-Encoder  & 71.0 & 30.1 & 41.9 & 45.9 & 60.6 & 70.3 & 71.9 & 48.7\\
C1 & A0: PointNet $\rightarrow$ 3DGCN-Encoder &- &{28.4} & {36.0} &{43.4} &{-} &{-} &68.0 &47.7 \\ 

\midrule
\textbf{D0} & A0 + prior in ST branch & {77.1} & {45.8} & {48.0} & {54.6} & {63.8} & {72.5} & {77.9} & {59.2}\\

\midrule
\textbf{E0} & D0: PointNet $\rightarrow$ HS-layer+LATs & 79.4 & 51.0 & 52.4 & 58.6 & 69.4 & 77.7 & 80.4 & 62.4\\ 
E1 & B0: No LAT on input points  & 76.1 & 39.3 & 46.6 & 53.0 & 65.4& 74.8 & 78.0 & 58.2 \\ 
E2 & B0: No LATs on features  & 78.5 & 48.8 & 47.4 & 53.0 & 67.4 & 75.0 & 80.4 & 57.4\\ 
E3 & B0: No LAT on the rotation feature  & \textbf{79.8} & 50.6 & 50.4 & 56.2 & 68.6  & 76.3 &80.2 & 60.8 \\ 
\midrule
F0 & E0+ Global Concatenation Fusion &{77.7} &{48.4} & {47.8} & {54.5} & {67.1} & {75.2} & {80.1} &{59.4} \\
\midrule
G0 & B0: No HS-layer in Matrix Net  & 77.8 & 50.2 & 54.1 & 60.1 & 70.5 & 78.0 & 81.2 & 63.6\\
\bottomrule
\end{tabular}
}

\vspace{-10px}
\end{center}
\end{table*}

\begin{table}[htbp]

\centering

\caption{\centering\textbf{Comparison with other methods on CAMERA25.} }
\vspace{-4mm}
\caption*{\footnotesize Higher score means better performance. Overall best results are in bold. {SPD$^*$} is the implementation results from CATRE, which is similar to the original SPD results.}
\vspace{-3mm}
\label{tbl:sota_perform_camera25}

\resizebox{1\linewidth}{!}
{\begin{tabular}{@{}l|c|cccc@{}}
\toprule
Method &  $\text{IoU}_{75}$ & $5^\circ 2$cm & $5^\circ 5$cm & $10^\circ 2$cm & $10^\circ 5$cm  \\ 

\midrule
\midrule

NOCS \cite{NOCS_cvpr_2019}    & 37.0 & 32.3 & 40.9 & 48.2 & 64.6 \\
DualPoseNet \cite{DualPoseNet_iccv_2021} &  71.7 & 64.7 & 70.7 & 77.2 & 84.7 \\
CR-Net \cite{CR-Net_iros_2021}   & 75.0 & 72.0 & 76.4 & 81.0 & 87.7 \\
SGPA \cite{SGPA_iccv_2021}   & 69.1 & 70.7 & 74.5 & 82.7 & 88.4 \\
SAR-Net \cite{SAR-Net_Lin_2022_CVPR} & 62.6 & 66.7 & 70.9 & 75.3 & 80.3 \\
SSP-Pose \cite{SSP-Pose_IROS_22} & - & 64.7 & 75.5 & - & 87.4 \\
RBP-Pose \cite{RBP-Pose_ECCV_2022} & - & {73.5} & 79.6 & {82.1} & {89.5}\\
GPV-Pose \cite{GPV-Pose_cvpr_2022} & - & 72.1 & 79.1 & - & 89.0\\
HS-Pose \cite{HSPose_2023_CVPR}  & - & 73.3 & {80.5} & 80.4 & 89.4\\ 
SPD$^*$ \cite{SPD_eccv_2020}   & 46.9 & 54.1 & 58.8 & 73.9 & 82.1 \\ 
\midrule
SPD$^*$$+$CATRE \cite{CATRE_eccv_2022} &  {76.1} & {75.4} & 80.3 & {83.3} & 89.3 \\
SPD$^*$$+$\textbf{Ours} ($2\%$) &  \underline{{77.5}} & \underline{{75.4}} &\underline{{81.1}} &\underline{{83.4}} &\underline{{90.0}} \\
SPD$^*$$+$\textbf{Ours}&  \textbf{{79.2}} & \textbf{{77.9}} &\textbf{{84.0}} &\textbf{{83.8}} &\textbf{{90.5}} \\

\bottomrule
\end{tabular}}

\end{table}
\section{Performance Comparion on CAMERA25}
Table~\ref{tbl:sota_perform_camera25} compares the accuracy of our method with the state-of-the-arts. 
As discussed in Sec.~\ref{sec:mini_dataset_camera25}, our performance stabilizes when using 4\% of the full train set. Therefore, we present the results obtained with this training size.
As shown in Table~\ref{tbl:sota_perform_camera25}, we greatly enhanced the performance of SPD, resulting in a performance that outperformed state-of-the-art pose estimation methods. Specifically, we improved the performance of SPD~\cite{SPD_eccv_2020} on $\IoUSevenFive$ by {32.7\%}, $\fiveDfiveC$ by {25.2\%}, and $\fiveDtwoC$ by {23.8\%}. 
We also outperform the baseline CATRE on $\IoUSevenFive$ by $3.1\%$, $\fiveDfiveC$ by $3.7\%$, and $\fiveDtwoC$ by $2.5\%$. 
Additionally, we show our results trained using 5k images (2\%) of the train set, which already outperforms the state-of-the-art methods.

\section{Per-category Performance}
\subsection{CAMERA25.}
We present our per-category object pose refinement performance in Table~\ref{tbl:per_cat_camera25}. We use SPD~\cite{SPD_eccv_2020} as the initial estimation method and report the performance after 4 refinement iterations. We show that our method largely improved the initial performance.
\begin{table*}[htbp]
\begin{center}

\caption{\textbf{Per-category results of our method on CAMERA25 dataset.} }
\label{tbl:per_cat_camera25}
\vspace{-3mm}
\resizebox{0.9\linewidth}{!}
{%
\footnotesize
\begin{tabular}{@{}c|c|cc|ccccc|cc@{}}
\toprule

Method & Category  & $\text{IoU}_{50}$ & $\text{IoU}_{75}$ & $5^\circ 2\text{cm}$ & $5^\circ 5\text{cm}$ & $10^\circ 2\text{cm}$ & $10^\circ 5\text{cm}$ & $10^\circ 10\text{cm}$ & $5^\circ $  & $2$cm   \\ 
\midrule
\midrule            
SPD               & bottle & 88.9 & 64.5 & 63.8 & 82.8 & 69.2 & 92.4 & 97.3 & 86.8 & 69.8 \\
SPD+\textbf{Ours} & bottle & 89.4 & 73.8 & 73.8 & 94.2 & 74.2 & 95.1 & 99.4&  98.4 & 74.2 \\
\midrule            
SPD               & bowl   & 95.9 & 80.6 & 83.4 & 83.7 & 95.8 & 96.3 & 96.3 & 83.7 & 99.2 \\
SPD+\textbf{Ours} & bowl & 96.0 & 94.7 & 97.9 & 98.2 & 99.5 & 99.8 & 99.8 & 98.2 & 99.6 \\
\midrule            
SPD               & camera & 61.9 & 4.7  & 27.3 & 29.3 & 72.9 & 78.6 & 78.6 & 29.5 & 89.8 \\
SPD+\textbf{Ours} & camera   & 81.6 & 67.7 & 83.1 & 87.2 & 90.8 & 95.2 & 95.2& 87.2 & 93.9 \\
\midrule            
SPD               & can    & 90.2 & 87.2 & 98.1 & 98.2 & 99.4 & 99.6 & 99.6 & 98.2 & 99.6 \\
SPD+\textbf{Ours} & can & 90.3 & 89.8 & 99.9 & 100.0 & 99.9 & 100.0 & 100.0 & 100.0 & 99.9 \\
\midrule
SPD               & laptop & 93.3 & 17.7 & 35.0 & 41.9 & 61.0 & 80.5 & 84.5 & 43.7 & 65.5 \\
SPD+\textbf{Ours} & laptop & 95.3 & 81.3 & 74.0 & 85.5 & 77.4 & 91.8 & 95.8 & 89.1 & 77.9 \\
\midrule
SPD               & mug    & 82.7 & 24.1 & 15.5 & 15.5 & 44.1 & 44.1 & 44.1 & 15.9 & 99.6 \\
SPD+\textbf{Ours} &  mug   & 89.8 & 67.7 & 39.0 & 39.0 & 61.0 & 61.0 & 61.0 & 39.4 & 99.9\\
\bottomrule
\end{tabular}
}
\vspace{-15px}
\end{center}
\end{table*}



\subsection{REAL275.}
We present the per-category object pose refinement results in Table~\ref{tbl:per_cat_real275}. We use SPD~\cite{SPD_eccv_2020} as the initial estimation method and report the performance after 4 refinement iterations. We show that our method largely improved the initial performance.
\begin{table*}[htbp]
\begin{center}

\caption{\textbf{Per-category results of our method on REAL275 dataset.} }
\label{tbl:per_cat_real275}
\vspace{-3mm}
\resizebox{0.9\linewidth}{!}
{%
\footnotesize
\begin{tabular}{@{}c|c|cc|ccccc|cc@{}}
\toprule

Method & Category  & $\text{IoU}_{50}$ & $\text{IoU}_{75}$ & $5^\circ 2\text{cm}$ & $5^\circ 5\text{cm}$ & $10^\circ 2\text{cm}$ & $10^\circ 5\text{cm}$ & $10^\circ 10\text{cm}$ & $5^\circ $  & $2$cm   \\ 
\midrule
\midrule            
SPD               & bottle & 49.9 & 13.1 & 21.6 & 23.2 & 69.4 & 76.0 & 87.1 & 35.9 & 80.7 \\
SPD+\textbf{Ours} & bottle & 49.8  & 36.2 & 64.8 & 68.0 & 82.5 & 88.6 & 100.0& 82.5 & 89.1 \\
\midrule            
SPD               & bowl   & 100.0& 77.1 & 50.5 & 54.0 & 75.8 & 80.3 & 80.3 & 54.0 & 94.7 \\
SPD+\textbf{Ours} & bowl   & 100.0& 91.9 & 91.2 & 95.6 & 95.4 & 100.0& 100.0& 95.7 & 95.4 \\
\midrule            
SPD               & camera & 43.4 & 3.4  & 0.0  & 0.0  & 0.2  & 0.2  & 0.2  & 0.0  & 34.8 \\
SPD+\textbf{Ours} & camera & 78.4 & 12.4 & 2.1  & 2.1  & 17.9 & 18.8 & 18.9 & 2.2  & 58.3 \\
\midrule            
SPD               & can    & 70.0 & 29.8 & 37.9 & 42.7 & 80.4 & 91.6 & 91.6 & 45.5 & 87.1 \\
SPD+\textbf{Ours} & can    & 70.3 & 36.7 & 75.6 & 78.6 & 96.0 & 99.9 & 99.9 & 80.7 & 96.0 \\
\midrule
SPD               & laptop & 82.0 & 35.5 & 4.6  & 7.0  & 24.5 & 65.3 & 65.9 & 7.1  & 29.1 \\
SPD+\textbf{Ours} & laptop & 80.8 & 73.9 & 67.6 & 91.8 & 68.9 & 94.4 & 95.6 & 92.5 & 69.3 \\
\midrule
SPD               & mug    & 66.5 & 8.7  & 0.3  & 0.3  & 10.3 & 10.4 & 10.4 & 0.3  & 85.2 \\
SPD+\textbf{Ours} &  mug   & 96.2 & 59.5 & 24.8 & 25.9 & 70.7 & 74.8 & 74.8 & 25.9 & 89.9 \\
\bottomrule
\end{tabular}
}
\vspace{-15px}
\end{center}
\end{table*}

\section{Additional Qualitative Results}
\begin{figure*}[h]
	\centering
	\begin{minipage}{0.10\linewidth}
	    scene 1-1
	\end{minipage}
 	\begin{minipage}{0.25\linewidth}
		\centering
		\includegraphics[width=\linewidth, page=1]{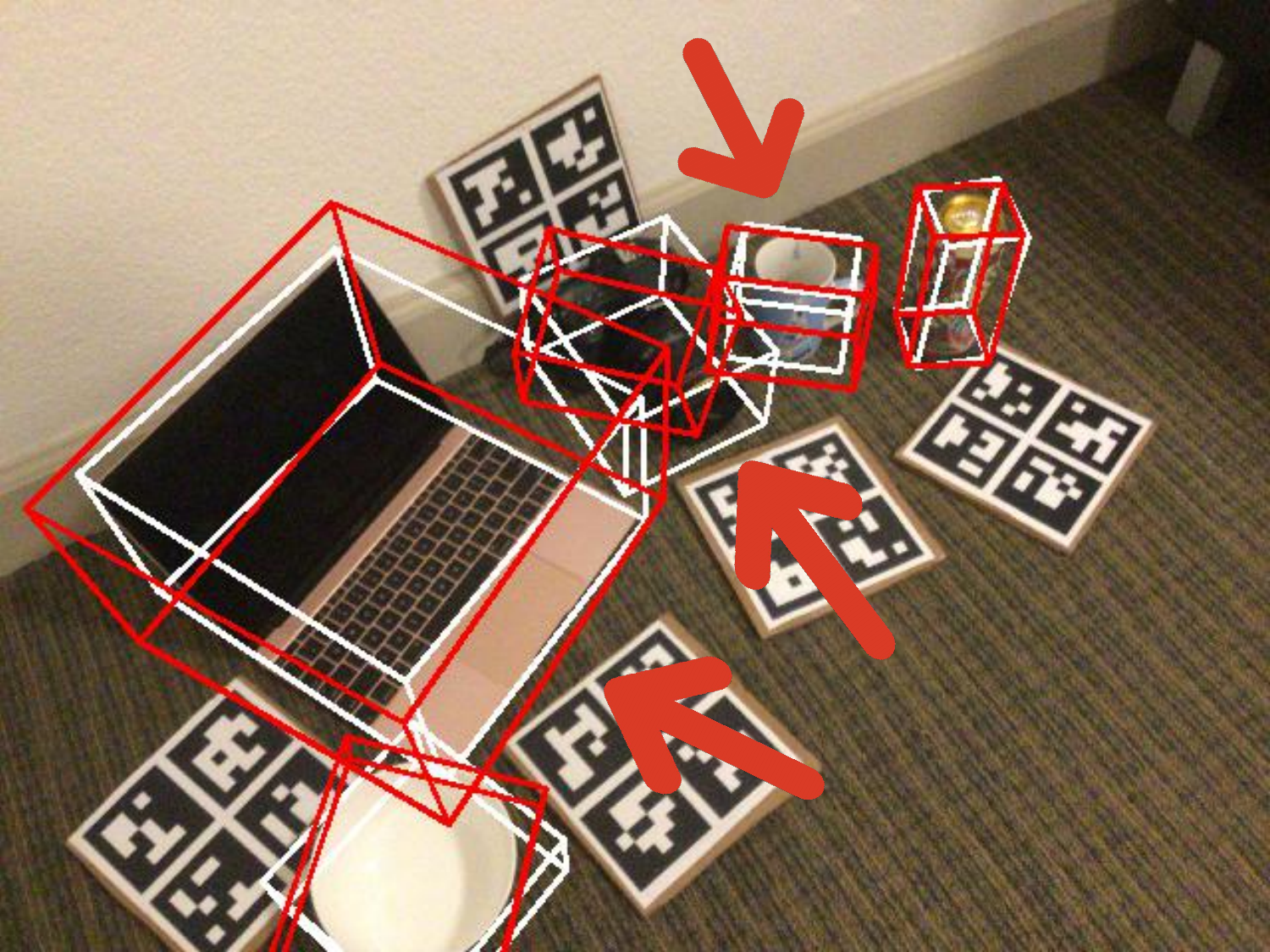}
	\end{minipage}
 	\begin{minipage}{0.25\linewidth}
		\centering
		\includegraphics[width=\linewidth, page=2]{fig/qualitative/scene1/1_288.pdf}
	\end{minipage}
 	\begin{minipage}{0.25\linewidth}
		\centering
		\includegraphics[width=\linewidth, page=3]{fig/qualitative/scene1/1_288.pdf}
	\end{minipage}

	\begin{minipage}{0.10\linewidth}
	    scene 1-2
	\end{minipage}
 	\begin{minipage}{0.25\linewidth}
		\centering
		\includegraphics[width=\linewidth, page=1]{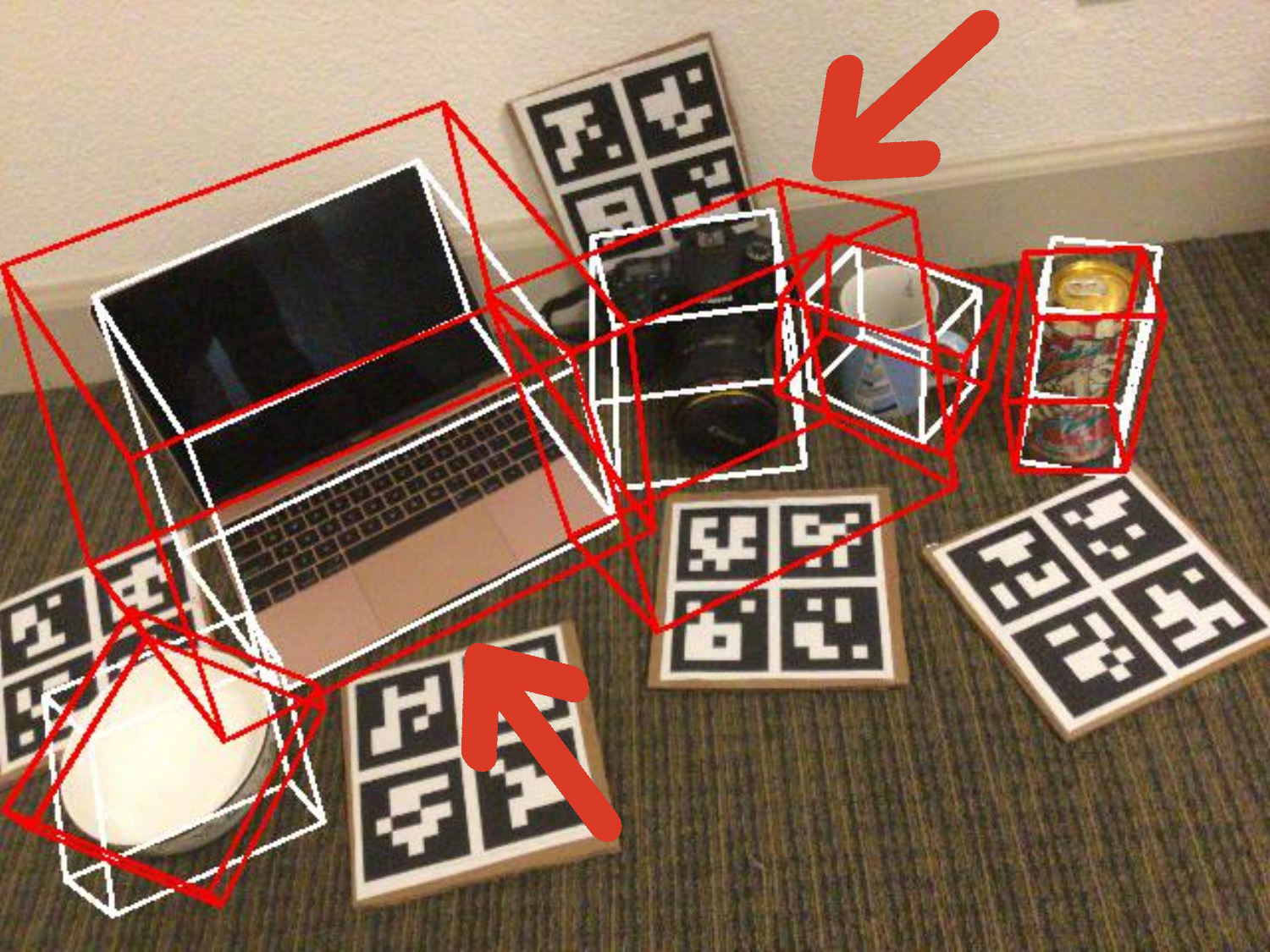}
	\end{minipage}
 	\begin{minipage}{0.25\linewidth}
		\centering
		\includegraphics[width=\linewidth, page=2]{fig/qualitative/scene1/1_345.pdf}
	\end{minipage}
 	\begin{minipage}{0.25\linewidth}
		\centering
		\includegraphics[width=\linewidth, page=3]{fig/qualitative/scene1/1_345.pdf}
	\end{minipage}

	\begin{minipage}{0.10\linewidth}
	    scene 2-1
	\end{minipage}
 	\begin{minipage}{0.25\linewidth}
		\centering
		\includegraphics[width=\linewidth, page=1]{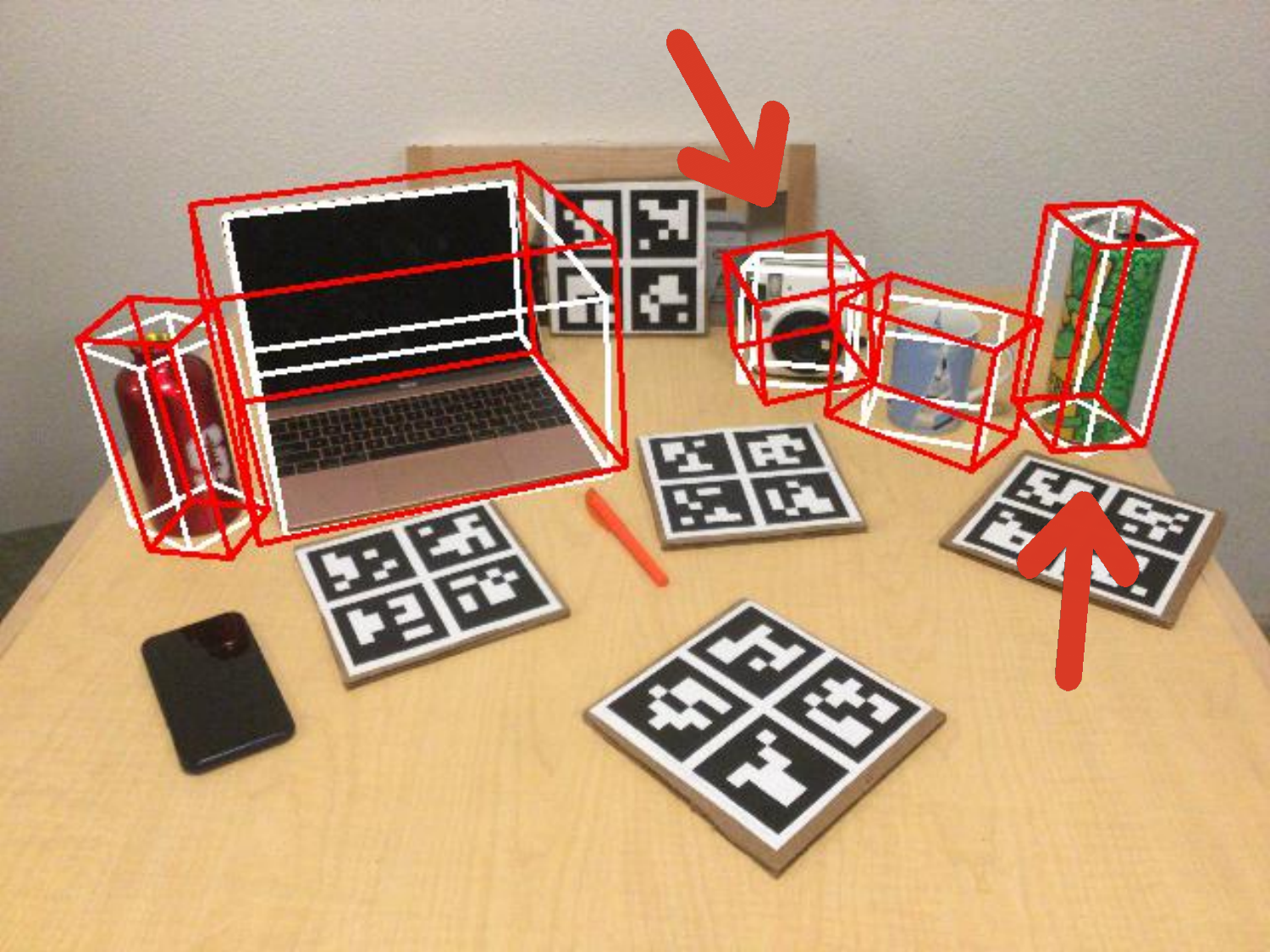}
	\end{minipage}
 	\begin{minipage}{0.25\linewidth}
		\centering
		\includegraphics[width=\linewidth, page=2]{fig/qualitative/scene2/2_007.pdf}
	\end{minipage}
 	\begin{minipage}{0.25\linewidth}
		\centering
		\includegraphics[width=\linewidth, page=3]{fig/qualitative/scene2/2_007.pdf}
	\end{minipage}

	\begin{minipage}{0.10\linewidth}
	    scene 2-2
	\end{minipage}
 	\begin{minipage}{0.25\linewidth}
		\centering
		\includegraphics[width=\linewidth, page=1]{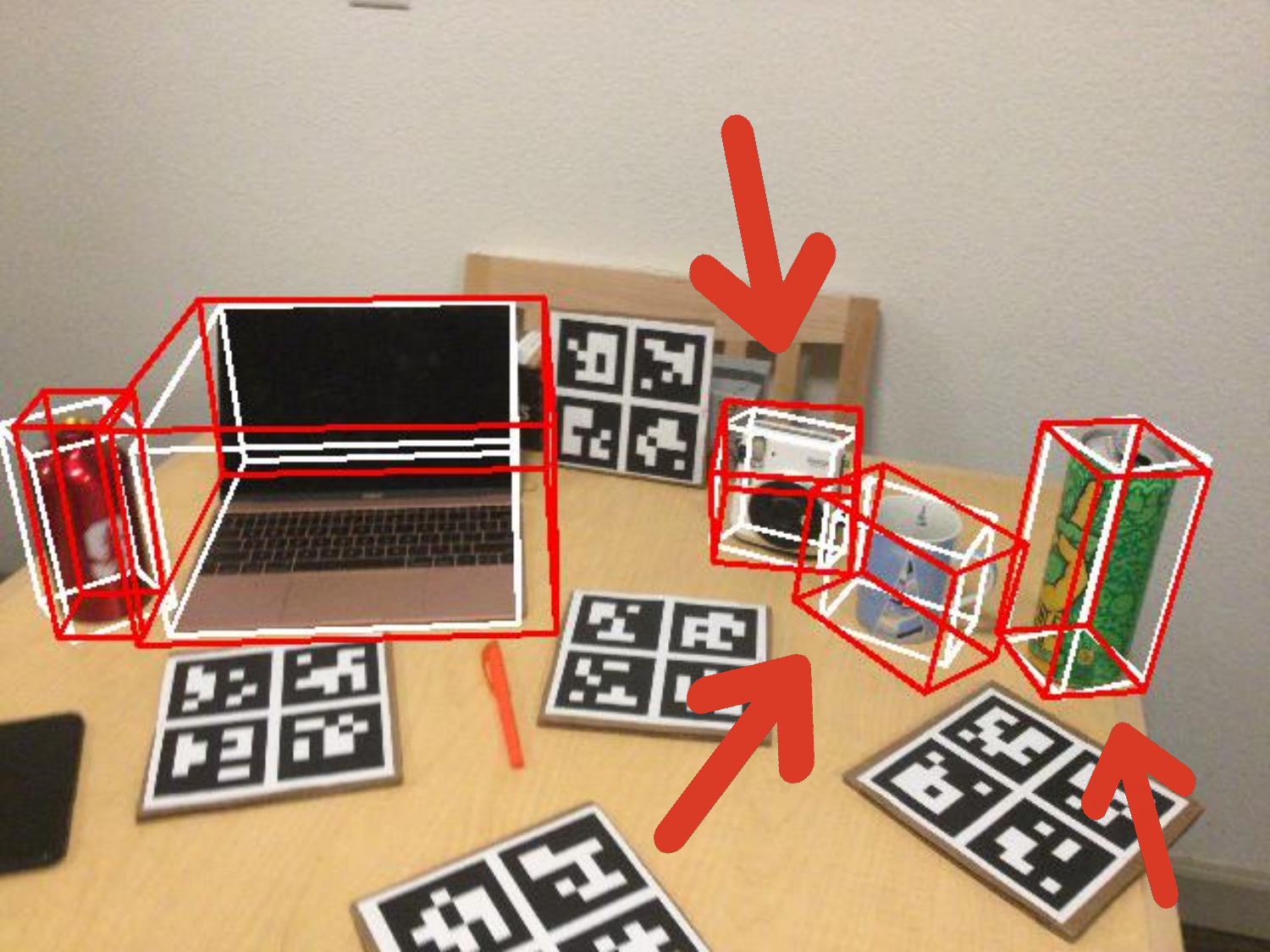}
	\end{minipage}
 	\begin{minipage}{0.25\linewidth}
		\centering
		\includegraphics[width=\linewidth, page=2]{fig/qualitative/scene2/2_310.pdf}
	\end{minipage}
 	\begin{minipage}{0.25\linewidth}
		\centering
		\includegraphics[width=\linewidth, page=3]{fig/qualitative/scene2/2_310.pdf}
	\end{minipage}
 
	\begin{minipage}{0.10\linewidth}
	    scene 3-1
	\end{minipage}
 	\begin{minipage}{0.25\linewidth}
		\centering
		\includegraphics[width=\linewidth, page=1]{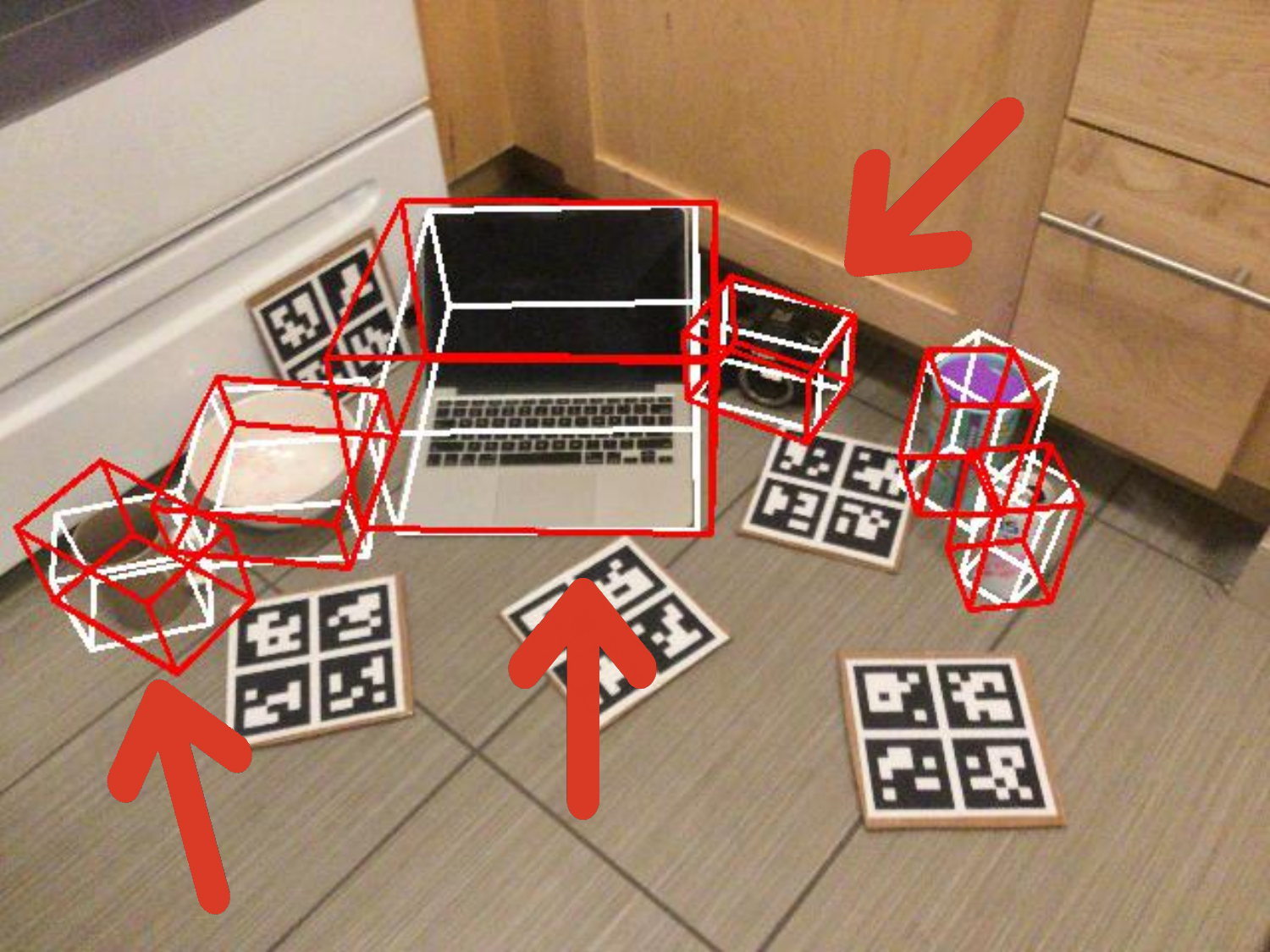}
	\end{minipage}
 	\begin{minipage}{0.25\linewidth}
		\centering
		\includegraphics[width=\linewidth, page=2]{fig/qualitative/scene3/3_297.pdf}
	\end{minipage}
 	\begin{minipage}{0.25\linewidth}
		\centering
		\includegraphics[width=\linewidth, page=3]{fig/qualitative/scene3/3_297.pdf}
	\end{minipage}
 
 	\begin{minipage}{0.10\linewidth}
	    scene 3-2
	\end{minipage}
 	\begin{minipage}{0.25\linewidth}
		\centering
		\includegraphics[width=\linewidth, page=1]{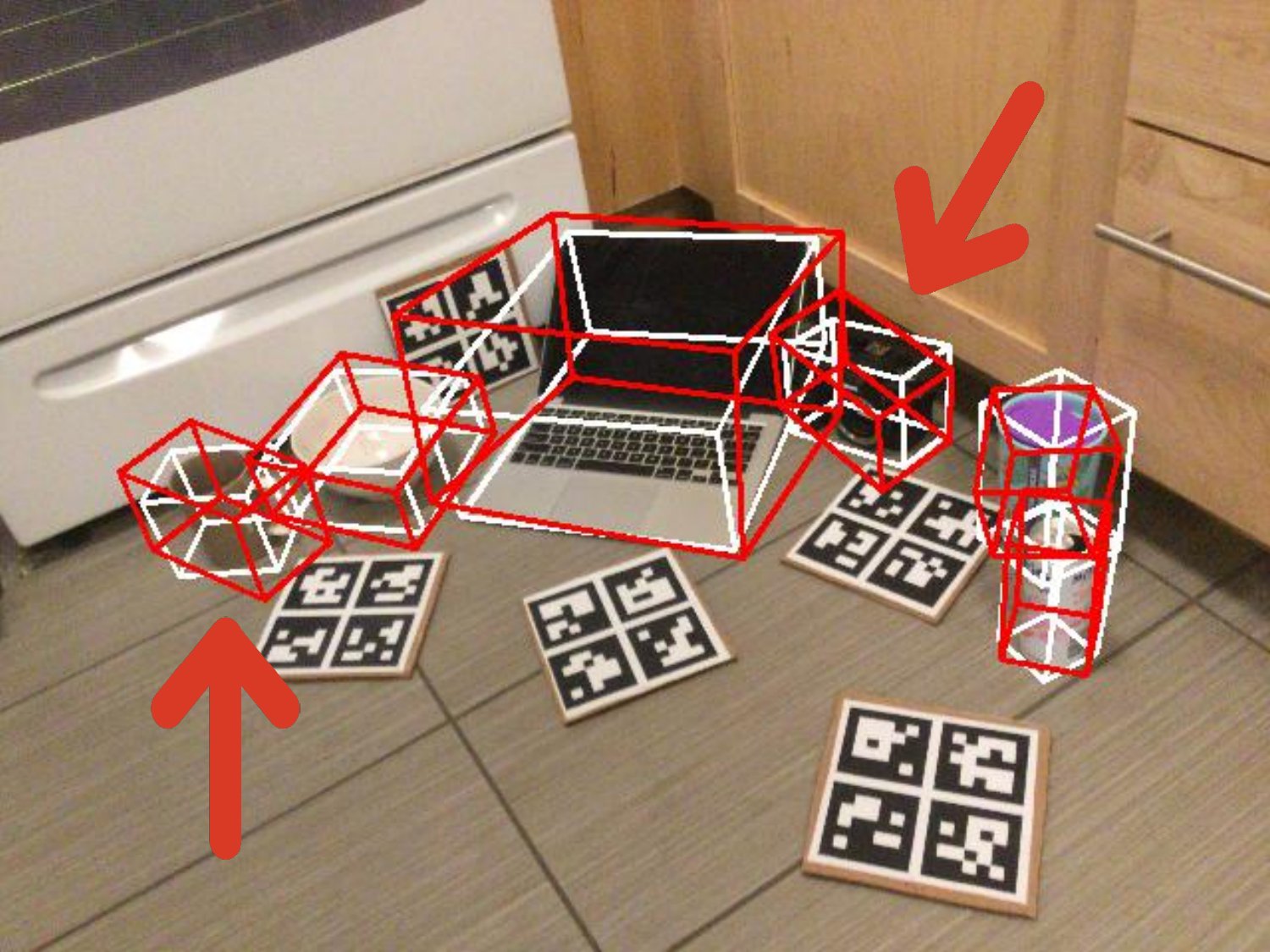}
	\end{minipage}
 	\begin{minipage}{0.25\linewidth}
		\centering
		\includegraphics[width=\linewidth, page=2]{fig/qualitative/scene3/3_429.pdf}
	\end{minipage}
 	\begin{minipage}{0.25\linewidth}
		\centering
		\includegraphics[width=\linewidth, page=3]{fig/qualitative/scene3/3_429.pdf}
	\end{minipage}

	\begin{minipage}{0.10\linewidth}
	{\color{white} scene 7}
	\end{minipage}
	\begin{minipage}[htbp]{0.25\linewidth}
	    \vspace{1mm}
	    \centering
		SPD
	\end{minipage}
	\begin{minipage}[htbp]{0.25\linewidth}
	    \vspace{1mm}
	    \centering
		SPD + CATRE
	\end{minipage}
	\begin{minipage}[htbp]{0.25\linewidth}
	    \vspace{1mm}
	    \centering
		SPD + Ours
	\end{minipage}

        \vspace{-2mm}
	\caption{\footnotesize \textbf{More qualitative comparison between the proposed method (column \#3) and the baseline method (column \#2) use the SPD (column \#1) as the initial estimation.} We choose two instances from each scene in REAL275 dataset. We show the ground truth with white lines. Note that the estimated rotations of symmetric objects (\eg bowl, bottle, and can) are considered correct if the symmetry axis is aligned.}
\label{fig:qualitative_supp_a}
\end{figure*}

\begin{figure*}[h]
	\centering
	\begin{minipage}{0.10\linewidth}
	    scene 4-1
	\end{minipage}
 	\begin{minipage}{0.25\linewidth}
		\centering
		\includegraphics[width=\linewidth, page=1]{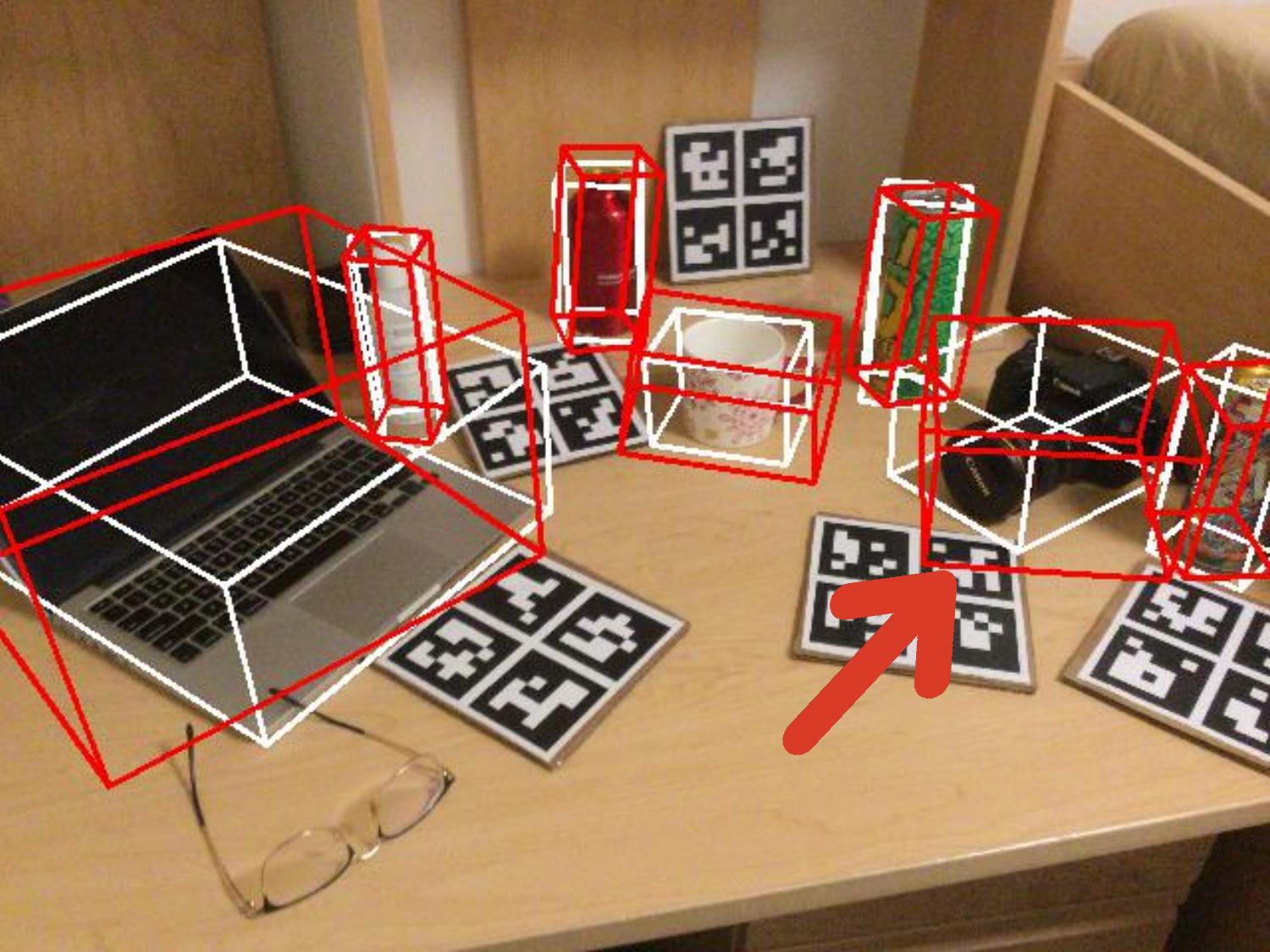}
	\end{minipage}
 	\begin{minipage}{0.25\linewidth}
		\centering
		\includegraphics[width=\linewidth, page=2]{fig/qualitative/scene4/4_133.pdf}
	\end{minipage}
 	\begin{minipage}{0.25\linewidth}
		\centering
		\includegraphics[width=\linewidth, page=3]{fig/qualitative/scene4/4_133.pdf}
	\end{minipage}
	
	\begin{minipage}{0.10\linewidth}
	    scene 4-2
	\end{minipage}
 	\begin{minipage}{0.25\linewidth}
		\centering
		\includegraphics[width=\linewidth, page=1]{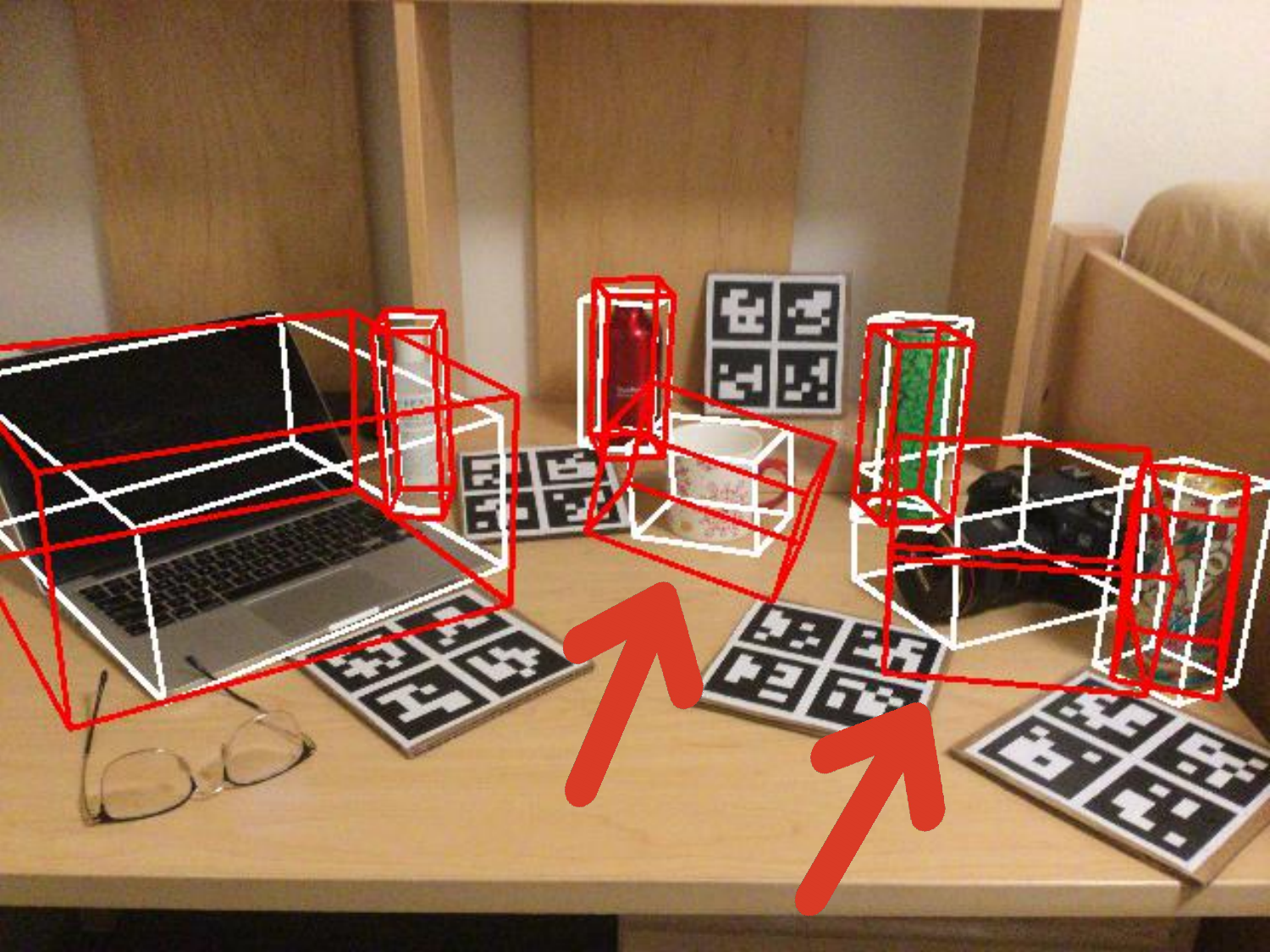}
	\end{minipage}
 	\begin{minipage}{0.25\linewidth}
		\centering
		\includegraphics[width=\linewidth, page=2]{fig/qualitative/scene4/4_367.pdf}
	\end{minipage}
 	\begin{minipage}{0.25\linewidth}
		\centering
		\includegraphics[width=\linewidth, page=3]{fig/qualitative/scene4/4_367.pdf}
	\end{minipage}

	\begin{minipage}{0.10\linewidth}
	    scene 5-1
	\end{minipage}
 	\begin{minipage}{0.25\linewidth}
		\centering
		\includegraphics[width=\linewidth, page=1]{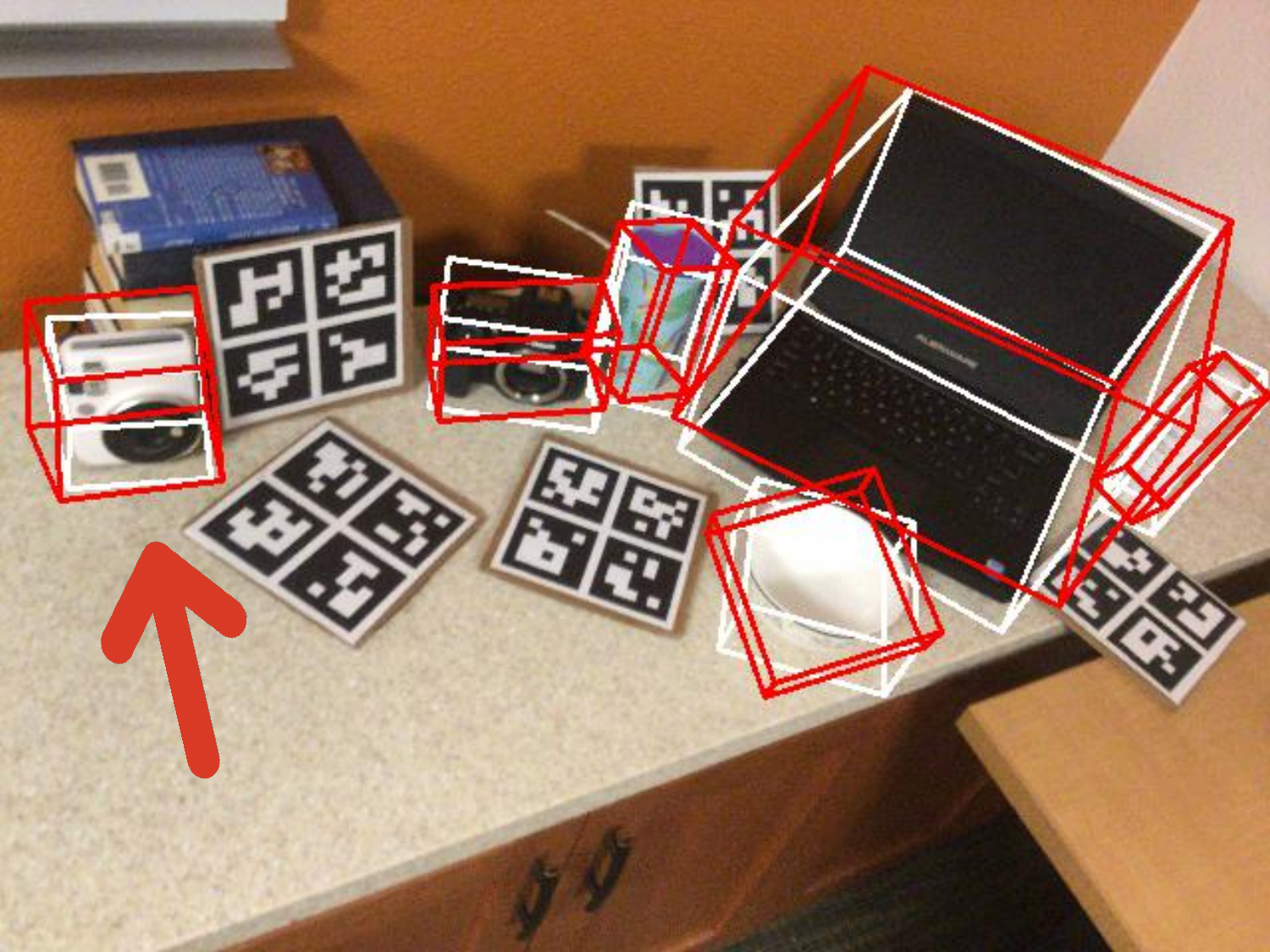}
	\end{minipage}
 	\begin{minipage}{0.25\linewidth}
		\centering
		\includegraphics[width=\linewidth, page=2]{fig/qualitative/scene5/5_232.pdf}
	\end{minipage}
 	\begin{minipage}{0.25\linewidth}
		\centering
		\includegraphics[width=\linewidth, page=3]{fig/qualitative/scene5/5_232.pdf}
	\end{minipage}

	\begin{minipage}{0.10\linewidth}
	    scene 5-2
	\end{minipage}
 	\begin{minipage}{0.25\linewidth}
		\centering
		\includegraphics[width=\linewidth, page=1]{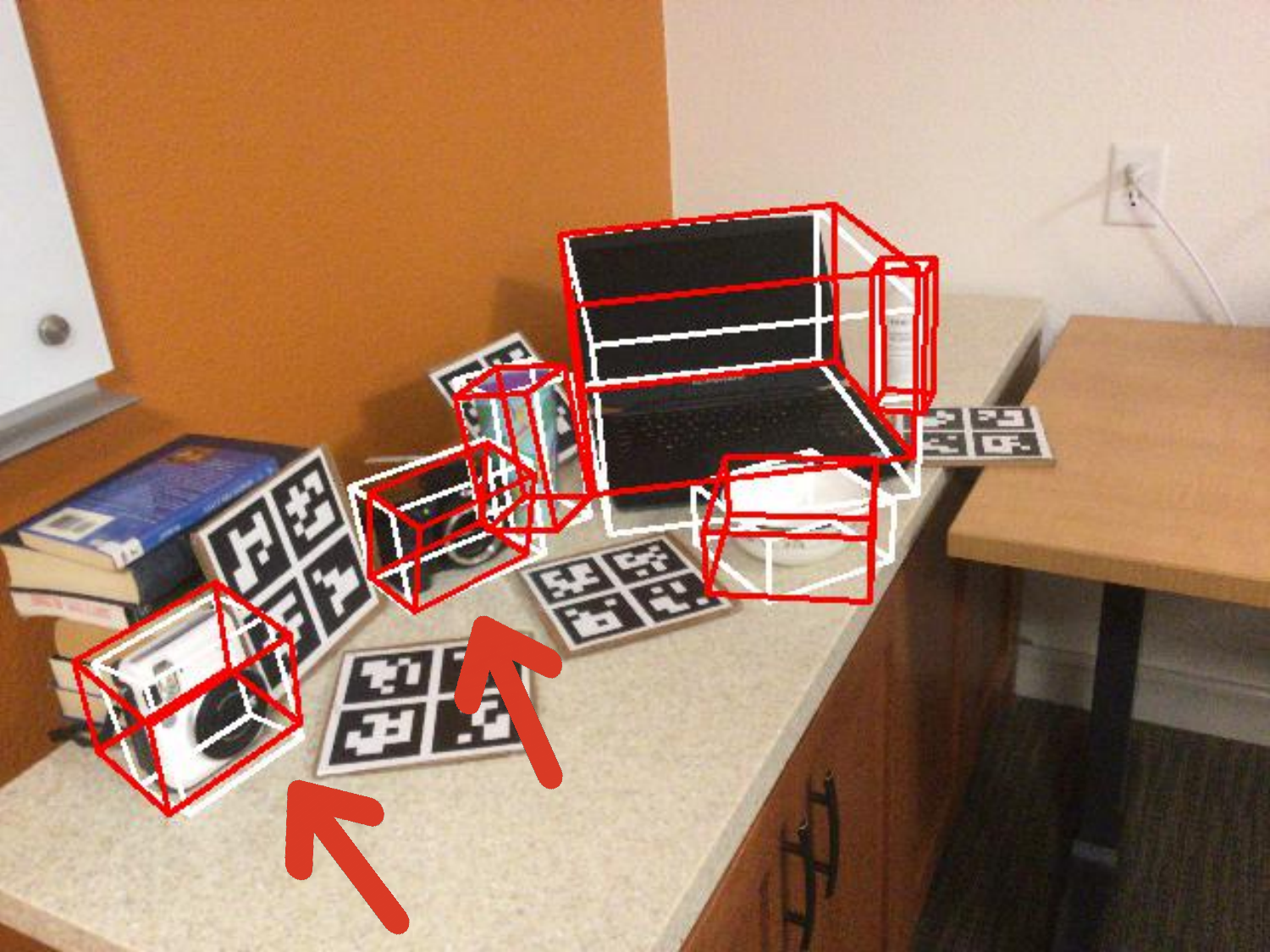}
	\end{minipage}
 	\begin{minipage}{0.25\linewidth}
		\centering
		\includegraphics[width=\linewidth, page=2]{fig/qualitative/scene5/5_317.pdf}
	\end{minipage}
 	\begin{minipage}{0.25\linewidth}
		\centering
		\includegraphics[width=\linewidth, page=3]{fig/qualitative/scene5/5_317.pdf}
	\end{minipage}
 
	\begin{minipage}{0.10\linewidth}
	    scene 6-1
	\end{minipage}
 	\begin{minipage}{0.25\linewidth}
		\centering
		\includegraphics[width=\linewidth, page=1]{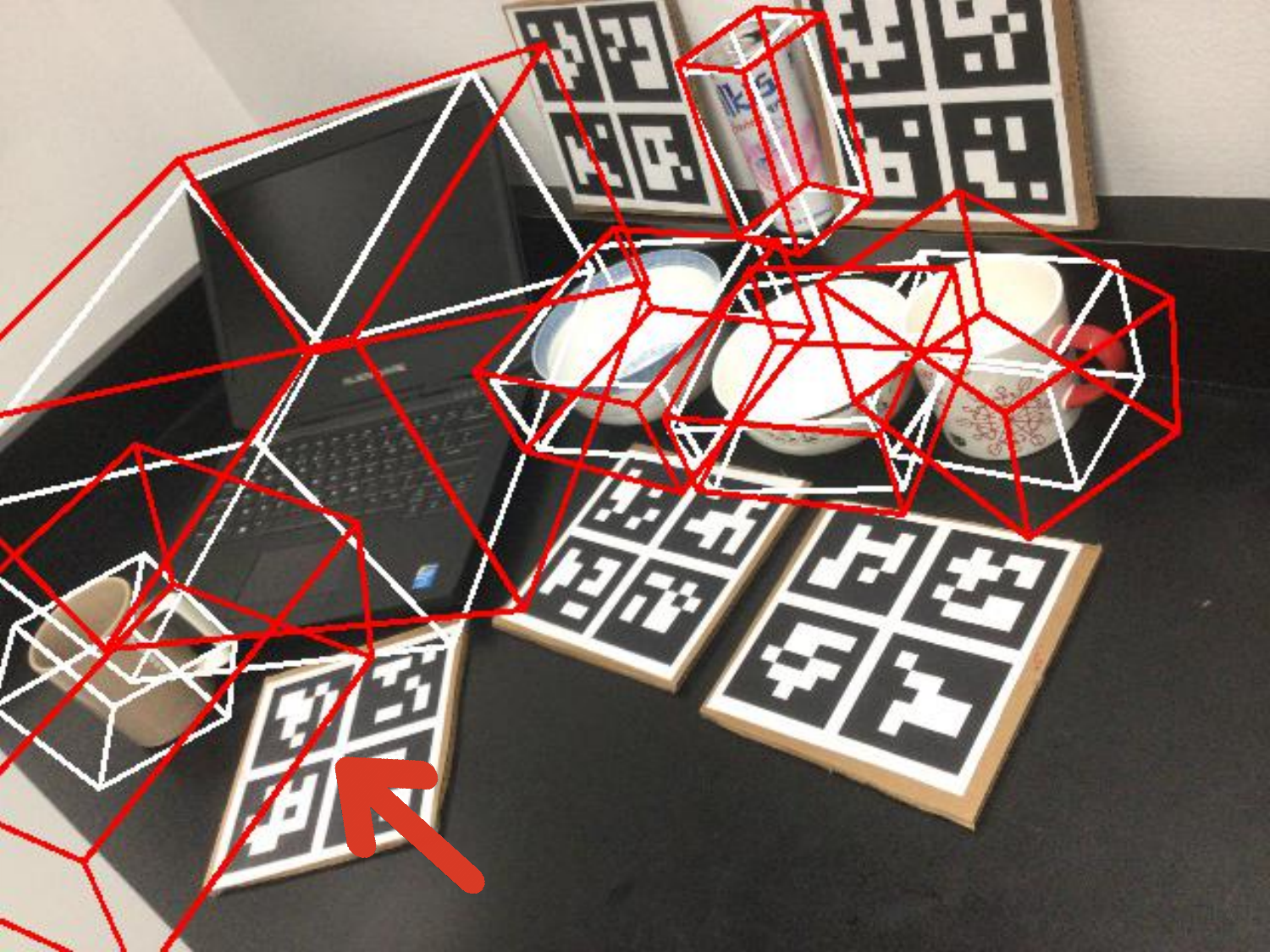}
	\end{minipage}
 	\begin{minipage}{0.25\linewidth}
		\centering
		\includegraphics[width=\linewidth, page=2]{fig/qualitative/scene6/6_489.pdf}
	\end{minipage}
 	\begin{minipage}{0.25\linewidth}
		\centering
		\includegraphics[width=\linewidth, page=3]{fig/qualitative/scene6/6_489.pdf}
	\end{minipage}
 
 	\begin{minipage}{0.10\linewidth}
	    scene 6-2
	\end{minipage}
 	\begin{minipage}{0.25\linewidth}
		\centering
		\includegraphics[width=\linewidth, page=1]{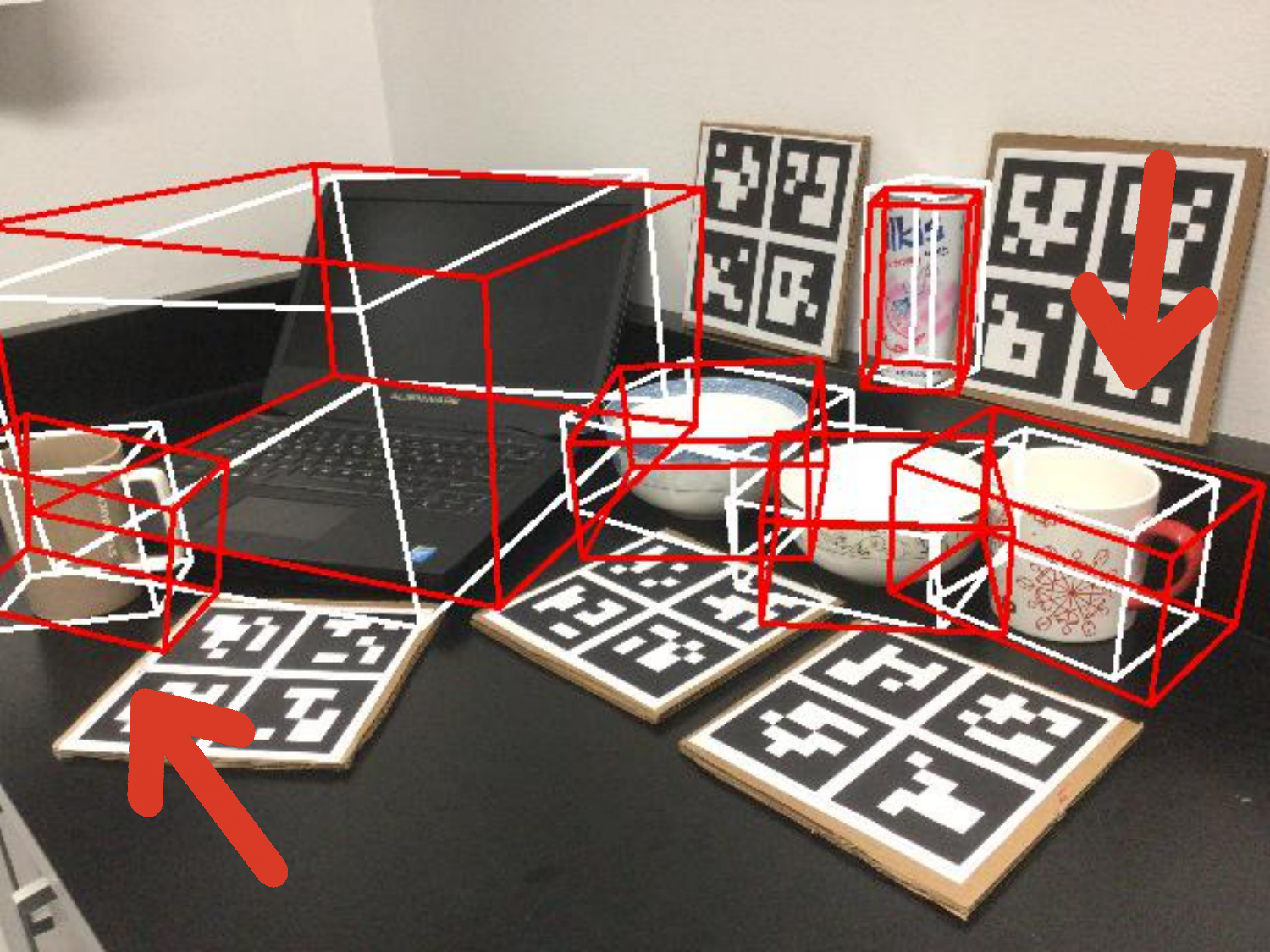}
	\end{minipage}
 	\begin{minipage}{0.25\linewidth}
		\centering
		\includegraphics[width=\linewidth, page=2]{fig/qualitative/scene6/6_569.pdf}
	\end{minipage}
 	\begin{minipage}{0.25\linewidth}
		\centering
		\includegraphics[width=\linewidth, page=3]{fig/qualitative/scene6/6_569.pdf}
	\end{minipage}

	\begin{minipage}{0.10\linewidth}
	{\color{white} scene 7}
	\end{minipage}
	\begin{minipage}[htbp]{0.25\linewidth}
	    \vspace{1mm}
	    \centering
		SPD
	\end{minipage}
	\begin{minipage}[htbp]{0.25\linewidth}
	    \vspace{1mm}
	    \centering
		SPD + CATRE
	\end{minipage}
	\begin{minipage}[htbp]{0.25\linewidth}
	    \vspace{1mm}
	    \centering
		SPD + Ours
	\end{minipage}

        \vspace{-2mm}
	\caption{\footnotesize \textbf{More qualitative comparison between the proposed method (column \#3) and the baseline method (column \#2) use the SPD (column \#1) as the initial estimation.} We choose two instances from each scene in REAL275 dataset. We show the ground truth with white lines. Note that the estimated rotations of symmetric objects (\eg bowl, bottle, and can) are considered correct if the symmetry axis is aligned.}
\label{fig:qualitative_supp_b}
\end{figure*}

We show additional qualitative results of our method test on different REAL275 test scenes in Fig.~\ref{fig:qualitative_supp_a} and Fig.~\ref{fig:qualitative_supp_b}. We highlight the performance differences with red arrows.

\clearpage
{
    \small
    \bibliographystyle{ieeenat_fullname}
    \bibliography{ref}
}
